\documentclass[10pt,twocolumn,letterpaper]{article}

\usepackage[pagenumbers]{iccv} % To force page numbers, e.g. for an arXiv version

\usepackage{multirow} % for merging rows
\usepackage{colortbl} % for cell background colors
\usepackage{tabularx} % for flexible-width tables
\usepackage{caption} % for table captions
\usepackage{graphicx} % for resizing tables or images
\usepackage{xcolor} % for custom colors
\usepackage[utf8]{inputenc} % allow utf-8 input
\usepackage[T1]{fontenc}    % use 8-bit T1 fonts
\usepackage{url}            % simple URL typesetting
\usepackage{booktabs}       % professional-quality tables
\usepackage{amsfonts}       % blackboard math symbols
\usepackage{nicefrac}       % compact symbols for 1/2, etc.
\usepackage{microtype}      % microtypography
\usepackage{xcolor}         % colors
\usepackage{tabularx}
\usepackage{subcaption}
\usepackage{multicol}
\usepackage{multirow}
\usepackage{natbib}
\usepackage{tabularray}
\usepackage{amsmath}
\usepackage{amssymb}
\usepackage{siunitx}
\usepackage{colortbl}
\usepackage{graphicx}
\usepackage{pifont}
\usepackage{lipsum}
\newcommand{\Xhline}[1]{\noalign{\hrule height #1}}
\usepackage[export]{adjustbox}
\usepackage{wrapfig}
\usepackage{svg}
\usepackage{amssymb}
\usepackage{sidecap}
\usepackage{caption}

\usepackage{algorithm}
\usepackage{algorithmicx}
\usepackage{amsmath}
\usepackage{xcolor}
\usepackage[linesnumbered,ruled,vlined,algo2e]{algorithm2e}

\usepackage{microtype}
\definecolor{cvprblue}{rgb}{0.21,0.49,0.74}
\definecolor{pltblue}{RGB}{174, 199, 232}
\definecolor{pltyellow}{HTML}{FFFACD} % a soft yellow
\definecolor{pltorange}{RGB}{255, 229, 204}
\definecolor{pltgreen}{RGB}{204, 229, 204}
\definecolor{pltred}{RGB}{229, 204, 204}
\definecolor{pltpurple}{RGB}{239, 218, 230}

\definecolor{tabblue}{HTML}{1f77b4}
\definecolor{taborange}{HTML}{ff7f0e}
\definecolor{tabgreen}{HTML}{2ca02c}
\definecolor{tabred}{HTML}{d62728}
\definecolor{tabpurple}{HTML}{9467bd}

\definecolor{cblue}{RGB}{173, 201, 233}
\definecolor{clblue}{RGB}{222, 234, 246}
\definecolor{corange}{RGB}{255, 152, 67}
\definecolor{lorgange}{RGB}{255, 221, 149}

\usepackage{pgfplots,gincltex}
\usepgfplotslibrary{groupplots}
\pgfplotsset{compat=1.6}
\usepackage{tikz}
\usetikzlibrary{arrows.meta}

\definecolor{iccvblue}{rgb}{0.21,0.49,0.74}
\usepackage[pagebackref,breaklinks,colorlinks,allcolors=iccvblue]{hyperref}
\usepackage[capitalize,noabbrev]{cleveref}

\def\paperID{766} % *** Enter the Paper ID here
\def\confName{ICCV}
\def\confYear{2025}

\title{Decouple to Reconstruct: High Quality UHD Restoration via Active Feature Disentanglement and Reversible Fusion}

\author{Yidi Liu$^{1,2}$, DongLi$^1$, Yuxin Ma$^1$, Jie Huang$^1$,Wenlong Zhang$^2$$^\dag$, Xueyang Fu$^1$$^\dag$, Zheng-jun Zha$^1$ \vspace{2pt} \\
$^1$University of Science and Technology of China, 
$^2$Shanghai AI Laboratory \\
{\tt\small liuyidi2023@mail.ustc.edu.cn, xyfu@ustc.edu.cn}\quad 
{\small $^\dag$ Corresponding Author} 
}

\begin{document}
\maketitle
\begin{abstract}

Ultra-high-definition (UHD) image restoration often faces computational bottlenecks and information loss due to its extremely high resolution. Existing studies based on Variational Autoencoders (VAE) improve efficiency by transferring the image restoration process from pixel space to latent space. However, degraded components are inherently coupled with background elements in degraded images, both information loss during compression and information gain during compensation remain uncontrollable. These lead to restored images often exhibiting image detail loss and incomplete degradation removal. To address this issue, we propose a Controlled Differential Disentangled VAE, which utilizes Hierarchical Contrastive Disentanglement Learning and an Orthogonal Gated Projection Module to guide the VAE to actively discard easily recoverable background information while encoding more difficult-to-recover degraded information into the latent space. Additionally, we design a  Complex Invertible Multiscale Fusion Network to handle background features, ensuring their consistency, and utilize a latent space restoration network to transform the degraded latent features, leading to more accurate restoration results. Extensive experimental results demonstrate that our method effectively alleviates the information loss problem in VAE models while ensuring computational efficiency, significantly improving the quality of UHD image restoration, and achieves state-of-the-art results in six UHD restoration tasks with only 1M parameters.

\end{abstract}    
\section{Introduction}
\label{sec:intro}

With the widespread adoption of 4K display devices, ultra-high-definition (UHD) image restoration has become a key challenge in the field of computer vision. Traditional pixel-level processing methods are limited by the high-resolution characteristics of UHD images, resulting in significant computational bottlenecks~\cite{Zamir2021Restormer,zou2024wavemambawaveletstatespace,wang2024uhdformer,wang2024ultrahighdefinitionrestorationnewbenchmarks,DreamUHD} 
% [xx].

\begin{figure}
    \centering
    \includegraphics[width=1\linewidth]{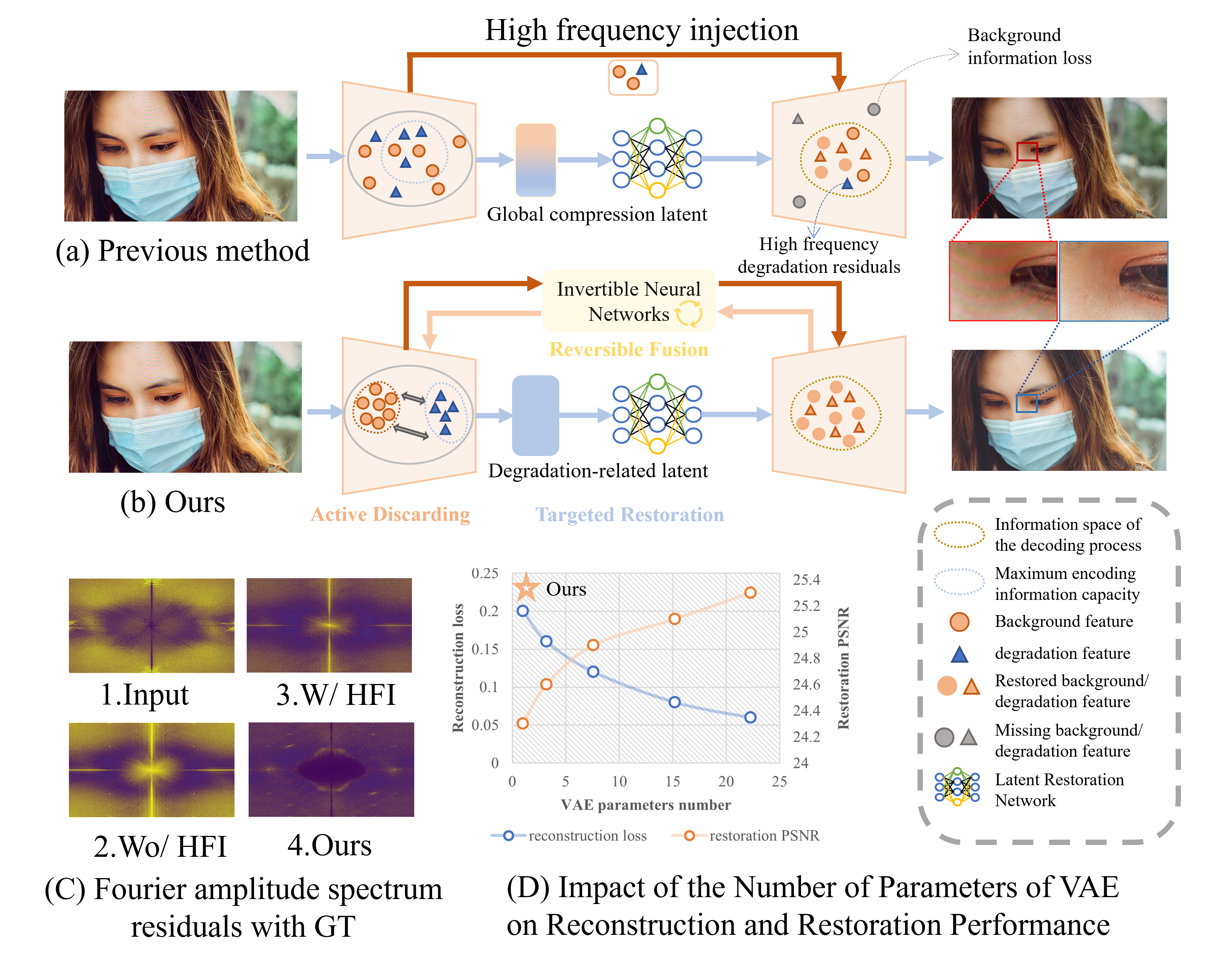}
    \captionsetup{skip=0.5pt}
    \caption{(a) Previous methods based on VAE suffered from irreversible background information loss due to global compression of the input image. Although additional High Frequency Injection (HFI) can partially mitigate the loss of high-frequency details, it also leads to residual high-frequency degradation.  (b) Our proposed CD²-VAE decouples background and degradation features and achieves a balance between background consistency and low residual degradation through a dual-path reconstruction structure.  (c) As the number of parameters in the VAE model increases, the reconstruction capability improves, leading to better restoration results. However, the excessive parameter count does not meet the lightweight requirements for UHD image restoration.}
    \vspace{-1.2em}
    \label{fig:intro}
\end{figure}

To alleviate the computational burden of UHD image processing, previous methods generally place the core restoration computation in the downsampled space of the original image. UHDFour~\cite{Li2023ICLR_uhdfour} and UHDformer~\cite{wang2024uhdformer} reduce the spatial dimension of the original input through PixelUnshuffle~\cite{shi2016real} and convolution operations, while LMAR~\cite{Empowering} designs a learning-based resampling operator. Clearly, the dimensionality reduction process leads to the loss of most of the information in the original image. Therefore, these methods often use a more lightweight network to process the image at its original resolution, thereby compensating for the information lost in the downsampling branch. DreamUHD~\cite{DreamUHD} develops a VAE framework that outperforms standard downsampling in compression efficiency. Its High-Frequency Injection strategy directly mitigates information loss during VAE encoding.

Since degraded components are inherently coupled with background elements in degraded images, both information loss during compression and information gain during compensation remain uncontrollable. These lead to restored images often exhibiting image detail loss and incomplete degradation removal. Consequently, UHD restoration still faces a significant challenge: \textit{the controllability of information compression and compensation}. This issue can be explained through three key aspects.
\textbf{(1) Representation bottleneck.} VAEs lose multi-frequency signals through non-discriminatory compression (~\cref{fig:intro}(c2)), despite their information retention capability. While enhancing VAE architecture lowers reconstruction errors (~\cref{fig:intro}(d)) and boosts restoration quality, the enlarged parameter size contradicts UHD restoration's efficiency requirements.
\textbf{(2) Entangled feature lost.} Compression eliminates both degraded/clean features due to their coupling (~\cref{fig:intro}(a)). This causes (i) Incomplete compensation (high-frequency recovered but low-frequency lost, ~\cref{fig:intro}(c3)) and (ii) Latent space overcrowding with clean features, diverting restoration efforts to irrelevant data. 
\textbf{(3) Uncertainty in the compensation process.} Since the compensation information is derived from the original image, which also contains tightly coupled degraded and background components (~\cref{fig:intro}(a)), the injected compensation features introduce degraded components, leading to degradation artifacts in the restoration results.

In this work, we aim to achieve controllable information loss compression and predictable information compensation through feature decoupling. 
We presents the Decoupled Dual-path UHD Restoration Network(D²R-UHDNet), a novel framework for UHD image restoration based on the Active Discarding-Targeted Restoration paradigm. Using Controlled Differential Disentangled Variational Autoencoder(CD²-VAE), the method decomposes the degraded input into two components: degradation-dominant latent space ($z_{\text{deg}}$) and background-dominant features (\(\{F_{\text{bg}}^l\}_{l=1}^L\)). This enables dual-path restoration: the Latent Restoration Network (LaReNet) maps the degraded latent $z_{\text{deg}}$ to the clean latent $z_{\text{clean}}$, while the Complex Invertible Multiscale Fusion Network (CIMF-Net) ensures information preservation and facilitates efficient background reconstruction through multi-scale fusion.

The CD²-VAE architecture establishes a joint optimization framework through a Hierarchical Contrastive Disentanglement Learning (Hi-CDL) strategy and an Orthogonal Gated Projection Module (OrthoGate). The Hi-CDL strategy employs multi-scale contrastive learning between clean and degraded features, enforcing the VAE to progressively disentangle background information while iteratively encoding retained features for degradation extraction. Concurrently, the OrthoGate module implements orthogonal projection within the Stiefel manifold space, mathematically guaranteeing minimal mutual information between decoupled feature subspaces~\cite{becigneulriemannian}. The synergistic effect of both components enables more precise feature disentanglement.

This work does not aim to “suppress” information loss by enhancing the representational capacity of the VAE but instead guides the VAE to actively discard easily recoverable information through feature disentanglement. Simultaneously, it effectively encodes the harder-to-recover information into the latent space. By increasing the density of degraded latent information while minimizing the compression loss of background features, our method enables a divide-and-conquer manner for background-dominant and degradation-dominant features.  This approach not only alleviates information loss but also maintains the lightweight nature of the network, meeting the practical requirements of UHD restoration.
Our contributions are summarized as follows:
\begin{itemize}
    \item We propose D²R-UHDNet for UHD image restoration, using the Active Discarding and Targeted Restoration paradigm to transform VAE's information loss into task-driven discarding, balancing degraded information removal and background preservation in a efficient manner.
    \item We propose CD²-VAE, which introduces Hi-CDL and OrthoGate. These components enable effective feature disentanglement of degraded inputs.
    % allowing for the separation of degraded information from background details.
    \item We propose CIMF-Net, which uses invertibility to ensure consistent information across multi-scale background-dominant features and employs complex operations for efficient cross-scale interaction. Our method achieves state-of-the-art results in six UHD restoration tasks with only 1M parameters.
\end{itemize}
\section{Related Works}
\label{sec:formatting}

\subsection{Variational Autoencoder.}

\begin{figure*}[t!]
	\centering
	\includegraphics[width=1\textwidth]{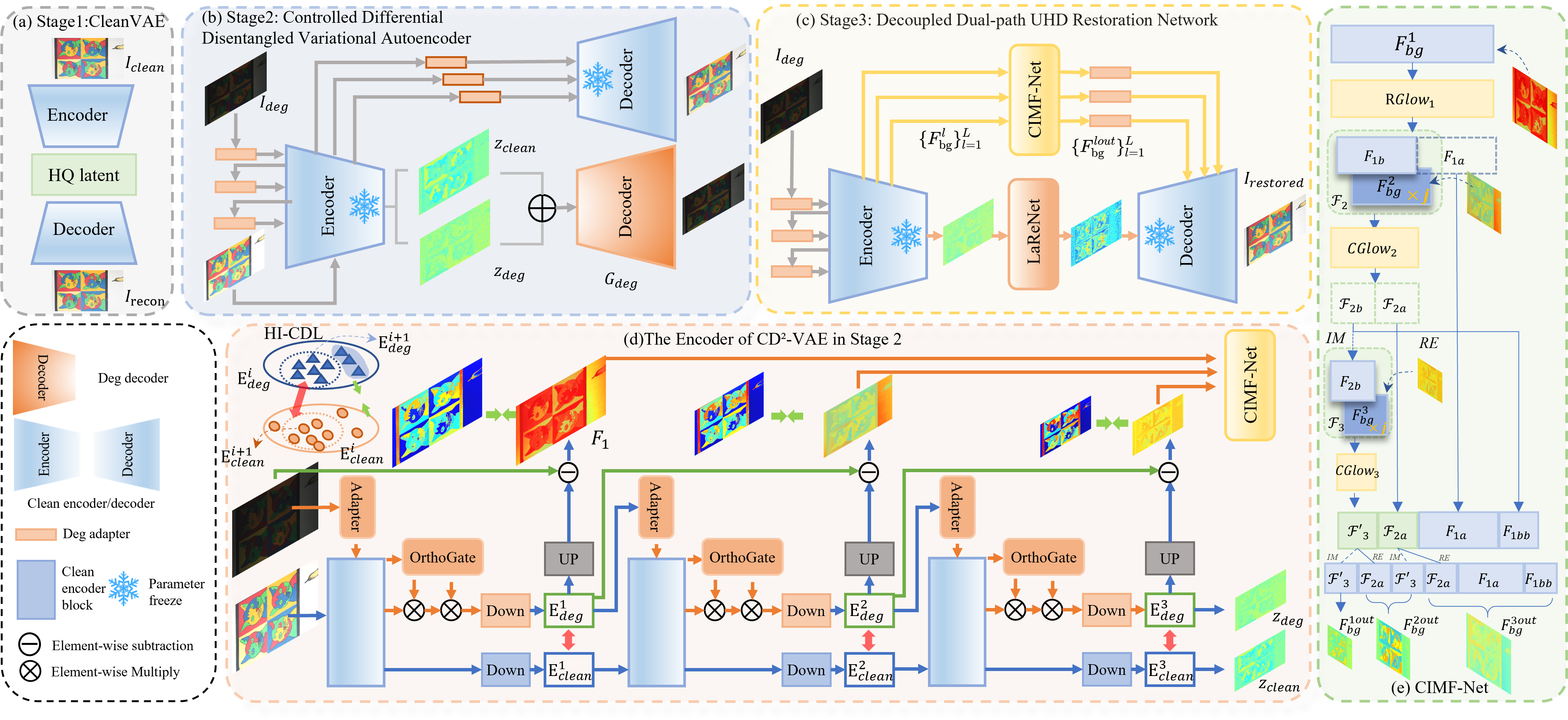} %
        \captionsetup{skip=0.5pt}
	\caption{The framework of our method: (a) In the first stage, CleanVAE is trained on clean images. (b) In the second stage, CD²-VAE is trained based on feature decoupling constraints. (c) In the third stage, D²R-UHDNet is constructed for image restoration using CD²-VAE.(d) Detailed illustration of the second stage CD²-VAE encoding process. (e) Structure of CIMF-Net. }
	\label{fig:framework}
\end{figure*}

Recently, the compact and informative latent space of VAEs has garnered increasing attention, leading to a wide range of applications. Latent Diffusion~\cite{LDM} transfers the diffusion process from pixel space to latent space, improving image generation efficiency. Old Photo Restoration (OPR)~\cite{Old_Photo} reduces the domain gap between synthetic and real degradation in the latent space, enhancing model generalization for real old photos.

DreamUHD is the first to use latent space compression for UHD image restoration, improving efficiency. To address high-frequency loss in VAE encoding, it introduces a high-frequency injection mechanism, reducing edge blurring. However, this injection also causes residual degradation and fails to suppress mid-to-low frequency information loss. To solve this, we propose a Controlled Differential Disentangled VAE that decouples features, making information loss controllable and balancing degradation removal with information retention.

\subsection{Ultra-High-Definition Restoration}

Recent advancements in image restoration technologies~\cite{Wang_2022_CVPR,Zamir2021Restormer,li2023efficient,jiang2024fast,cui2024revitalizing} have improved performance at traditional resolutions but struggle with computational efficiency and performance when applied to UHD images. To meet the increasing demand for UHD image processing, methods such as bilateral learning for dehazing~\cite{Zheng_uhd_CVPR21}, multi-scale networks for video deblurring~\cite{uhd_video_deblurring}, Fourier embedding for low-light enhancement~\cite{Li2023ICLR_uhdfour}, and wavelet-based Mamba networks~\cite{zou2024wavemambawaveletstatespace} have been proposed.

Recently, general architectures for UHD image restoration have rapidly developed, addressing multiple restoration problems with a unified structure. To mitigate the large computational cost of UHD image processing, UHDformer~\cite{wang2024uhdformer} constructs two branches for high and low resolutions using PixelShuffle, UHDDIP integrates gradient and normal priors through an external model~\cite{wang2024ultrahighdefinitionrestorationnewbenchmarks}, and DreamUHD~\cite{DreamUHD} introduces latent space modeling with VAE to enhance restoration efficiency. In this context, the prior in UHDDIP requires an additional model for computation, which makes it less efficient. In contrast, UHDformer and DreamUHD aim to allocate computationally intensive structures to the downsampled space while assigning lighter structures to the original space, thus improving efficiency. However, performing downsampling compression on all the information in the image leads to irreversible information loss, and the lightweight structures applied to the original space are often too simplistic, which can result in residual degradation in the restored image. In this paper, D²R-UHDNet introduces an Active Discarding and Targeted Restoration paradigm, which avoids the uncontrollable loss of information during downsampling, significantly enhancing the performance of UHD restoration tasks.

\section{Method}

\subsection{Overview}

This paper proposes the Decoupled Dual-path UHD Restoration Network (D²R-UHDNet), whose core objective is to achieve efficient and high-quality restoration of Ultra-High Definition (UHD) images through guided feature disentanglement and dual-path collaborative restoration.

Our overall framework is shown in ~\cref{fig:framework}. In the first stage, we train a CleanVAE on clean images for image reconstruction, as shown in ~\cref{fig:framework}(a). Next, we introduce feature decoupling constraints to separately reconstruct clean and degraded images, resulting in the trained CD²-VAE, as shown in ~\cref{fig:framework}(b). In the third stage, D²R-UHDNet is constructed based on CD²-VAE for UHD image restoration. The detailed encoding process of CD²-VAE and CIMF-Net for handling multi-scale background-dominated features in D²R-UHDNet are shown in ~\cref{fig:framework}(d) and ~\cref{fig:framework}(e), respectively. We will provide a detailed introduction to these processes in ~\cref{sec:CD²-VAE} and ~\cref{sec:CIMF-Net}.

\subsection{Controlled Differential Disentangled VAE}
\label{sec:CD²-VAE}

Unlike traditional VAEs, CD²-VAE introduces Hierarchical Contrastive Disentanglement Learning (Hi-CDL) and the Orthogonal Gated Projection Module (OrthoGate), transforming the encoding process into a controllable information filtering operation: background features that are easy to restore are discarded first, while degradation components requiring targeted restoration are carefully preserved.

Given the degraded Ultra-High Definition (UHD) image \(I_{\text{deg}} \in \mathbb{R}^{H \times W \times 3}\), CD²-VAE utilizes Hi-CDL and OrthoGate to progressively disentangle background features \(\{F_{\text{bg}}^l\}_{l=1}^L\), while retaining the degraded latent encoding \(z_{\text{deg}} \in \mathbb{R}^{\frac{H}{2^{L}} \times \frac{W}{2^{L}} \times 3}\).

% \subsubsection{Hierarchical Contrastive Disentanglement Learning}
\subsubsection{Hierarchical Contrastive Disentanglement}

To drive the progressive disentanglement of degraded and background features across multiple scales, this paper introduces Hierarchical Contrastive Disentanglement Learning (Hi-CDL). The core idea is to explicitly guide the encoder by comparing the cross-level similarity between degraded features and clean reference features, encouraging the active discard of background information irrelevant to degradation at each layer, while preserving the information relevant to the degradation.

For the $i$-th layer encoder, the input degraded and clean features are $E^{i-1}_{deg},E^{i-1}_{clean}\in  \mathbb{R}^{h \times w \times c} $, and the output of this layer is $E^{i}_{deg},E^{i}_{clean}\in  \mathbb{R}^{\frac{h}{2} \times \frac{w}{2} \times 2c} $, where features are progressively compressed during the encoding process. We aim for the CD²-VAE to progressively disentangle $E^{i-1}_{deg}$from background-irrelevant features during encoding and increase the similarity between $E^{i}_{deg}$ and $E^{i}_{clean}$. Simultaneously, to ensure that the information reduced from $E^{i-1}_{deg}$ to $E^{i}_{deg}$ is mainly background information, we aim to reduce the feature discrepancy between $E^{i-1}_{deg}$ and $E^{i}_{deg}$, while enhancing their similarity to the corresponding background features. This process can be expressed as follows:
\begin{equation}
\begin{split}
\mathcal{L}_{\text{contrast}}^{i} &= -\log \frac{\exp(s_{\text{pos}}^{i}/\tau_i)}{\exp(s_{\text{pos}}^{i}/\tau_i) + \exp(s_{\text{neg}}^{i}/\tau_i)+ 
\epsilon }, \\
s_{\text {neg }}^{i}&=\operatorname{sim}\left(E^{i}_{deg}, E^{i}_{clean}\right), \;  \\
s_{\text {pos }}^{i}&=\operatorname{sim}\left(E^{i-1}_{deg}-\text{UP}(E^{i}_{deg}), E^{i-1}_{clean}\right),
\end{split}
\end{equation}
where $\text{sim}(\cdot)$ denotes the cosine similarity calculation, and $\text{UP}(\cdot)$ represents the Pixel Shuffle and Channel Duplication operation, $\epsilon$ is a numerical stabilization term. $\tau_i$ is the temperature coefficient. As the layer  $i$ increases, we gradually reduce $\tau_i$ to strengthen the feature disentanglement, allowing the disentanglement process to shift from global information to fine-grained details.

\subsubsection{ Orthogonal Gated Projection Module}

\begin{figure}[t!]
	\centering
	\includegraphics[width=0.5\textwidth]{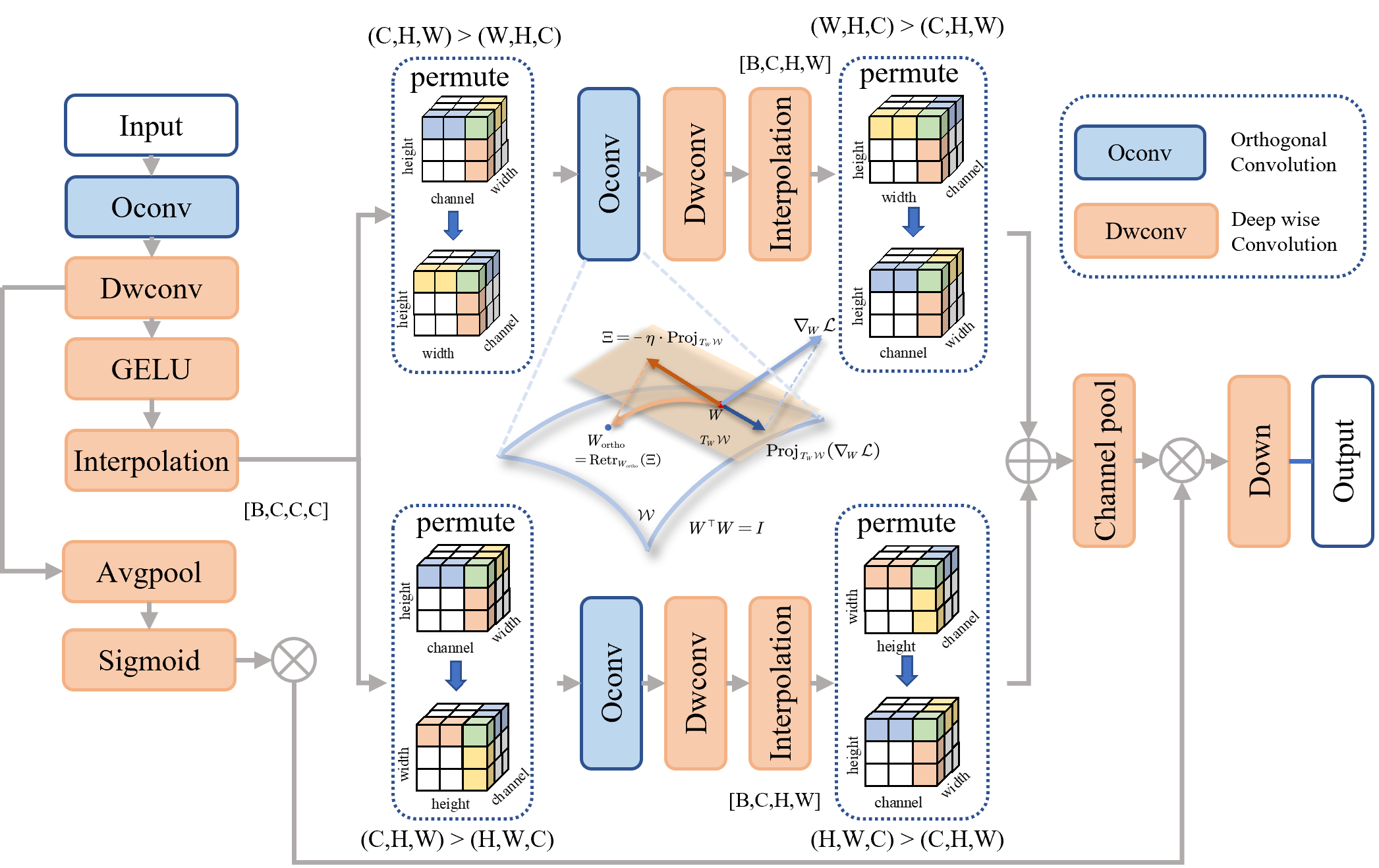} %
        \captionsetup{skip=0.5pt}
	\caption{Orthogonal Gated Projection Module.}
	\label{fig:framework}
\end{figure}

The Hi-CDL strategy explicitly constrains the disentanglement process of CD²-VAE by constructing a loss function. Additionally, we introduce the Orthogonal Gated Projection Module (OrthoGate) to control the flow of information during the encoding process.

Specifically, to further promote feature disentanglement, we construct a orthogonal pointwise convolution($\rm O_{conv}$) based on the Stiefel manifold constraint~\cite{becigneulriemannian}, with the convolution parameters denoted as $W_{\text{ortho}}$. Due to the Stiefel manifold constraint, this matrix satisfies the orthogonality condition:
\begin{equation}
     \mathcal{W} = \left\{ W \in \mathbb{R}^{C \times C} \mid W^\top W = I \right\} 
\end{equation}

The parameters of this convolution are optimized using a Riemannian gradient-based update strategy to satisfy the Stiefel manifold constraint:
\begin{equation}
    \begin{split}
        W_{\text{ortho}} \leftarrow \operatorname{Retr}_{W_{\text{ortho}}} \left( -\eta  \cdot \operatorname{Proj}_{T_W\mathcal{W}}  \left( \nabla_W \mathcal{L} \right) \right),\\
         \operatorname{Proj}_{T_{W}\mathcal{W}}(\nabla_{W}\mathcal{L}) = \nabla_{W}\mathcal{L} - W \cdot \operatorname{sym}(W^{\top}\nabla_{W}\mathcal{L}),\\ 
    \end{split}
\end{equation}

where, $\operatorname{Retr}$ denotes the retraction, $\operatorname{Proj}_{T_W\mathcal{W}}(\cdot)$ is the tangent space projection operator, $\mathcal{L}$ is the loss function, and $\operatorname{sym}(A) = \frac{A + A^{\top}}{2}$ represents the symmetrization of the matrix $A$.

First, we construct a channel disentanglement gating based on $\rm O_{conv}$. At the $i$-th layer, $E^{i}_{deg}$, after passing through the Encoder block, becomes $E^{i'}_{deg}\in  \mathbb{R}^{h\times w \times 2c}$ before the downsampling step. We first use $\rm O_{conv}$ to reduce the inter-channel feature correlation, and then apply depthwise convolution to enable spatial dimension interaction while preserving the channel disentanglement. Next, we apply global average pooling followed by the Sigmoid activation function to obtain the channel gating factor $C_{gate}\in  \mathbb{R}^{1 \times 1 \times 2c} $. This factor is then dot-multiplied with $E^{i'}_{deg}$ to obtain the output of the channel disentanglement gating, $E^{i^{c}}_{deg}$. This process can be expressed as follows:
\begin{equation}
    \begin{split}
    E_{deg}^{i^{1}} &= \operatorname{D_{conv}}\operatorname{O_{conv}}(E_{deg}^{i'}) \\
        C_{gate} &= \mathcal{S}(\text{GAP}(E_{deg}^{i'})))), \\
        E_{deg}^{i^c} &= C_{gate} \odot E_{deg}^{i'},
    \end{split}
\end{equation}

where, $\mathcal{S}(\cdot)$ denotes the Sigmoid activation function, $\operatorname{GAP}(\cdot)$ represents global average pooling, $\operatorname{D_{conv}}(\cdot)$ refers to depthwise convolution, and $\odot$ denotes the element-wise dot product operation.

After disentangling the features in the channel dimension, we further disentangle the spatial dimensions. First, we interpolate $E_{deg}^{i^{1}}$ so that its spatial dimensions match the channel dimension, resulting in $E_{deg}^{i^{2}} \in \mathbb{R}^{c \times c \times c}$. Then, we perform a permute operation on $E_{deg}^{i^{2}}$, moving the $h$ and $w$ dimensions to the channel dimension, followed by orthogonal convolution and depthwise convolution. After that, we apply interpolation and permute operations to restore the features to their original dimensions, obtaining $E_{deg}^{i^{h2}} \in \mathbb{R}^{h \times w \times c}$.This process can be expressed as follows:
\begin{equation}
    \begin{split}
        E_{deg}^{i^{h1}} &= \operatorname{D_{conv}}(\operatorname{O_{conv}}(\underset{\operatorname{h,w,c}{\rightarrow}  \operatorname{c,h,w}}{\operatorname{Permute
        }}(\mathcal{I}_{p}(E_{deg}^{i^{1}},  \mathbb{R}^{c \times c \times c}))),\\
        E_{deg}^{i^{h2}} &=  \underset{\operatorname{h,w,c}{\rightarrow}  \operatorname{c,h,w}}{\operatorname{Permute
        }}(\mathcal{I}_{p}(E_{deg}^{i^{h1}},  \mathbb{R}^{h \times w \times c}))),
    \end{split}
\end{equation}
where \(\mathcal{I}_{p}(\cdot, \mathbb{R}^{m \times n \times c})\) is an interpolation operation that adjusts the input dimensions to \(\mathbb{R}^{m \times n \times c}\), and \(\underset{\operatorname{h,w,c} \rightarrow \operatorname{c,h,w}}{\operatorname{Permute}}\) denotes the permutation operation that rearranges the dimensions from \((h, w, c)\) to \((c, h, w)\). Similarly, by applying \(\underset{\operatorname{h,w,c} \rightarrow \operatorname{c,w,h}}{\operatorname{Permute}}\), we obtain \(E_{deg}^{i^{w2}}\) through the same process.

Subsequently, we add \(E_{deg}^{i^{h2}}\) and \(E_{deg}^{i^{w2}}\), and perform channel pooling to obtain the spatial gating factor \(S_{gate} \in \mathbb{R}^{h \times w \times 1}\). This factor is then element-wise multiplied with \(E_{deg}^{i^c}\) to yield the spatial disentanglement gating output \(E_{deg}^{i^p}\). Finally, \(E_{deg}^{i^p}\) undergoes downsampling to produce the output of the Encoder for this layer, \(E_{deg}^{i+1}\). This process can be expressed as follows:
\begin{equation}
    \begin{split}
        S_{gate} &= \mathcal{S}(\mathcal{C}(E_{deg}^{i^{h2}}+E_{deg}^{i^{w2}})),\\
    E_{deg}^{i+1}   &= \operatorname{Down}(E_{deg}^{i^c} \odot S_{gate})  
    \end{split}
\end{equation}

% \vspace{-1em}
where $\mathcal{C}(\cdot)$ represents the channel average pooling operation, and $\operatorname{Down}$ denotes the downsampling module in the encoder block.

\subsection{Complex Invertible Multiscale Fusion Network}
\label{sec:CIMF-Net}

The multi-scale background features \(\{F_{\text{bg}}^l\}\) are input into CIMF-Net, where cross-scale information fusion is achieved through invertible computation in the complex domain. The degraded latent encoding \(z_{\text{deg}}\) is input into a lightweight latent-space restoration network, mapping it to the clean distribution. The restored background features \(\{F_{\text{bg}}^{lout}\}\) and the clean encoding \(z_{\text{clean}}\) are then input into the decoder \(G_{\text{dec}}\) to generate the final restored image \(I_{\text{restored}}\).

By combining complex domain representation with invertible computation, CIMF-Net enables efficient cross-scale interaction and consistent information transfer. We employ GLOW~\cite{kingma2018glow} as the invertible module within CIMF-Net.

For multi-scale background features \(\{F_{\text{bg}}^l\}_{l=1}^L\), we begin with \(F_{\text{bg}}^1 \in \mathbb{R}^{h \times w \times c}\), which is processed by the real-valued invertible network Real-GLOW to produce the output \(F_{\text{bg}}^{1'} \in \mathbb{R}^{h \times w \times c}\). Next, we apply PixelShuffle and channel splitting to \(F_{\text{bg}}^{1'} \in \mathbb{R}^{h \times w \times c}\), obtaining \(F_{\text{bg}}^{1^1}, F_{\text{bg}}^{1^2} \in \mathbb{R}^{\frac{h}{2} \times \frac{w}{2c} \times c}\).

Then, \(F_{\text{bg}}^{1^1}\) is output directly, while \(F_{\text{bg}}^{1^2}\) and \(F_{\text{bg}}^2\) are combined as the real and imaginary parts, respectively, and passed into the next layer of the complex invertible network Complex-GLOW to produce the output \(F_{\text{bg}}^{1^{2'}} + F_{\text{bg}}^{2'} j\). We then apply PixelShuffle and channel splitting again, resulting in \(F_{\text{bg}}^{1^{21}} + F_{\text{bg}}^{2^1} j, F_{\text{bg}}^{1^{22}} + F_{\text{bg}}^{2^2} j \in \mathbb{R}^{\frac{h}{4} \times \frac{w}{4} \times 4c}\). Finally, \(F_{\text{bg}}^{2^2}\) and \(F_{\text{bg}}^3\) are combined as the real and imaginary parts, respectively, and passed into another layer of Complex-GLOW, resulting in \(F_{\text{bg}}^{2^{2'}} + F_{\text{bg}}^{3'} j\). After all features are processed, we assemble them into the final outputs $F_{\text{bg}}^{1out}$,$F_{\text{bg}}^{2out}$, $F_{\text{bg}}^{3out}$. This process can be expressed as~\cref{alg:cimf}, where \(\mathcal{PS}\) denotes the PixelUnshuffle and channel split operations, and $\mathcal{A}$ represents the feature recomposition operation, where Pixelshuffle and channel concatenation are employed to reconstruct the features at each scale into their original shape.

\begin{algorithm}[H]
\caption{Complex Invertible Multiscale Fusion (CIMF)}
\label{alg:cimf}
\SetAlgoLined
% \KwIn{Multi-scale features $\{F_l\}_{l=1}^L$}
% \KwOut{Fused feature $F_{\text{out}}$}
% \BlankLine
\textbf{Input:}{Multi-scale features $\{F_l\}_{l=1}^L$} \\
\textbf{Output:}{Fused feature $F_{\text{out}}$} \\
\textbf{Level 1 Processing:} \;
\BlankLine
$\begin{aligned}
    &F_1' \leftarrow \text{RealGLOW}(F_1); (F_{1a}, F_{1b}) \leftarrow \mathcal{PS}(F_1') \\
    &F_{\text{keep}} \leftarrow F_{\text{keep}} \cup \{F_{1a}\};F_{\text{pass}} \leftarrow F_{1b}
\end{aligned}$\;

\BlankLine

\textbf{Intermediate Levels:} \;
\For{$l \leftarrow 2$ \KwTo $L-1$}{
    $\begin{aligned}
        &\mathcal{F}_l \leftarrow \text{ComplexGLOW}(
            F_{\text{pass}}+j F_l) \quad 
            \textcolor{gray}{\triangleright\ \text{Complex fusion}} \\
        &(\mathcal{F}_{la}, \mathcal{F}_{lb}) \leftarrow \mathcal{PS}(\mathcal{F}_l) \\
        &F_{\text{keep}} \leftarrow F_{\text{keep}} \cup \{\mathcal{F}_{la},\text{Re}(\mathcal{F}_{lb})\} \\
        &F_{\text{pass}} \leftarrow \text{Im}(\mathcal{F}_{lb})
    \end{aligned}$\;
}
\BlankLine

\textbf{Final Level Processing ($l=L$):} \;
\BlankLine
$\begin{aligned}
    &\mathcal{F}_L \leftarrow \text{ComplexGLOW}(
        F_{\text{pass}}, F_L) \\
    &F_{\text{keep}} \leftarrow F_{\text{keep}} \cup \{
        \text{Re}(\mathcal{F}_L), \text{Im}(\mathcal{F}_L)\}
\end{aligned}$\;

\BlankLine

\textbf{Assembly:} \;
$\{F^{out}_{\text{l}}\}^{L}_{l=1} \leftarrow \mathcal{A}(F_{\text{keep}})$ \textcolor{gray}{$\triangleright$\ \text{ Recomposition}} \;

\Return{$\{F^{out}_{\text{l}}\}^{L}_{l=1}$}
\end{algorithm}

\(\operatorname{Complex-GLOW}\) consists primarily of the Complex Invertible 1×1 Convolution, Complex ActNorm, and Complex Affine Coupling. First, we construct the core operator for complex computation, the complex convolution. For a complex convolution kernel \(W_{\text{complex}} = W_{\text{re}} + jW_{\text{im}}\), the computation process is as follows:
\begin{equation}
    \begin{split}
        \operatorname{Re}(\mathcal{F}_{\text{out}}) &= (\operatorname{Re} (\mathcal{F}) * W_{\text{re}} - \operatorname{Im}(\mathcal{F}) * W_{\text{im}}) \\
        \operatorname{Im}(\mathcal{F}_{\text{out}}) &= (\operatorname{Re}(\mathcal{F}) * W_{\text{im}} + \operatorname{Im}(\mathcal{F}) * W_{\text{re}})
    \end{split}
\end{equation}

Based on the complex convolution kernel, we apply polar decomposition to impose a unitary matrix constraint on \( W_{\text{complex}} \), ensuring its invertibility:
\begin{equation}
    \begin{split}
        &W_{\text{complex}}^\dagger W_{\text{complex}} = I \quad \Rightarrow \quad |\det(W_{\text{complex}})| = 1, \\
        &W_{\text{complex}} = U \cdot \Sigma, \quad U^\dagger U = I, \; \Sigma \succeq 0,
    \end{split}
\end{equation}

The real and imaginary parts of the complex features are separately normalized using affine normalization:
\begin{equation}
    \begin{split}
        \operatorname{Re}(\mathcal{F}_{\text{out}}) &= \frac{\operatorname{Re}(\mathcal{F}) - \mu_{\text{re}}}{\sigma_{\text{re}}} \cdot s_{\text{re}} + b_{\text{re}} \\\operatorname{Im}(\mathcal{F}_{\text{out}}) &= \frac{\operatorname{Im}(\mathcal{F}) - \mu_{\text{im}}}{\sigma_{\text{im}}} \cdot s_{\text{im}} + b_{\text{im}}
    \end{split}
\end{equation}

where, \(\{\mu_{\text{re}}, \mu_{\text{im}}\}\) and \(\{\sigma_{\text{re}}, \sigma_{\text{im}}\}\) are the per-channel statistics, while \(\{s_{\text{re}}, s_{\text{im}}\}\) and \(\{b_{\text{re}}, b_{\text{im}}\}\) are the learnable parameters.

The input complex features \(\mathcal{F}\) are split into \(\mathcal{F}_1\) and \(\mathcal{F}_2\), and the scale \(\gamma\) and offset \(\beta\) are predicted using the complex neural network \(\operatorname{ComplexNN}\):
\begin{equation}
    \begin{split}
       \gamma_1 + j\gamma_2, \beta_1 + j\beta_2 = \text{ComplexNN}(\mathcal{F}_1), \\
       \mathcal{F}_2' = \mathcal{F}_2 \odot (\gamma_1 + j\gamma_2) + (\beta_1 + j\beta_2),
    \end{split}
\end{equation}
% 其中 \(\operatorname{ComplexNN}\)由复数卷积模块构成。
The \(\operatorname{ComplexNN}\) is composed of a complex convolution module. The output is \( \mathcal{F}_{\text{out}} = \operatorname{Concat}(\mathcal{F}_1, \mathcal{F}_2') \), and the inverse transformation can be explicitly computed.

\subsection{Latent Restoration Network }

For the degraded latent \(z_{\text{deg}} \in \mathbb{R}^{\frac{H}{2^{L}} \times \frac{W}{2^{L}} \times 3}\), we construct the Latent Restoration Network (LaReNet) to map the degraded latent \(z_{\text{deg}}\) to the clean latent \(z_{\text{clean}}\). This network can be constructed using a simple image restoration network module. This work uses SFHformer~\cite{jiang2024fast}, while other image restoration network modules are explored in the ablation study.
\section{Experiment}

\subsection{Experimental Setup}

For single degradation removal, we followed the dataset settings from DreamUHD \cite{DreamUHD}, using UHD-LL~\cite{Li2023ICLR_uhdfour} for low-light enhancement, UHD-Haze~\cite{Zheng_uhd_CVPR21} for dehazing, UHD-Blur~\cite{Deng_2021_ICCV_uhd_denlurring} for deblurring, and UHDM~\cite{yu2022towards} for de-moiréing. More detailed dataset settings and additional visual comparisons are provided in the supplementary material. FLOPs are computed with an input size of 256 × 256, and inference time is tested at 4K resolution on an RTX 3090.

\subsection{Comparison with State-of-the-Art Methods}

\begin{figure*}[t!]
	\centering
	\includegraphics[width=1\textwidth]{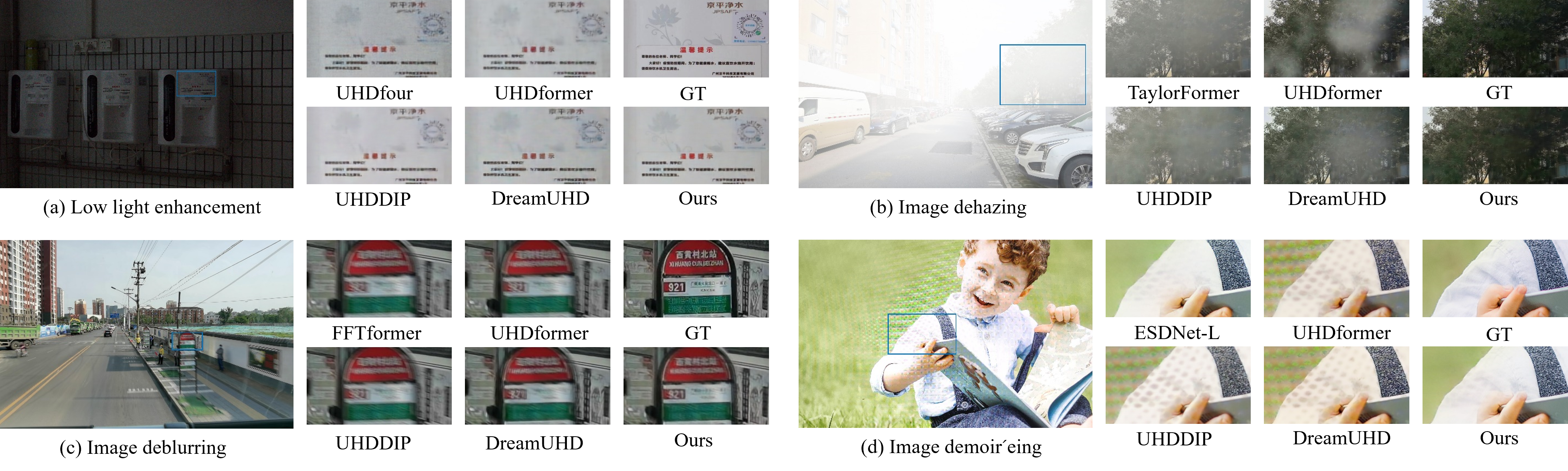} %
        % \captionsetup{skip=0.5pt}
	\caption{A comparison of visual results for four types of degradation removal with other state-of-the-art (SOTA) all-in-one methods.}
	\label{fig:visual_result}
\end{figure*}

\subsubsection{Low-Light Image Enhancement Results}

The results for Low-Light Image Enhancement are shown in ~\cref{fig:visual_result}(a). Compared to other methods, our approach preserves text details under extremely low light conditions, yielding clearer results. The quantitative comparison results are shown in ~\cref{tab:exp:low-light}, where our method achieves the best performance while maintaining lightweight design.

\subsubsection{Image Dehazing Results}

The visual comparison results for Image Dehazing are shown in ~\cref{fig:visual_result}(b). Compared to other methods, our approach results in less haze residue, yielding cleaner results. The quantitative comparison results, as shown in ~\cref{tab:exp:image dehazing}, demonstrate a 0.68dB PDNR improvement over the previous SOTA results, validating the effectiveness of our method.

\subsubsection{Image Deblurring Results}

The visual comparison results for Image Deblurring are shown in ~\cref{fig:visual_result}(c). Compared to other methods, our method better restores motion-induced blur artifacts. The quantitative comparison results, as shown in ~\cref{tab:exp:image deblurring}, demonstrate that our method achieves optimal performance.

\subsubsection{Image Demoiréing Results}

The visual comparison results for Image Demoiréing are shown in ~\cref{fig:visual_result}(b). Compared to other methods, our approach more thoroughly removes moiré patterns from the image. The quantitative comparison results, as shown in ~\cref{tab:exp:image deblurring}, demonstrate a 0.68dB PSNR performance improvement over previous methods.

\subsubsection{Four Degradations All in one Results}

The Active Discarding and Targeted Restoration paradigm proposed in this paper gradually decouples background-dominated information unrelated to degradation from the degraded images, making it naturally suited for the \textit{All in One} task, where a single model addresses multiple degradation scenarios. Without any additional design, as shown in ~\cref{tab:exp:4deg}, our method still achieves the best results for the UHD All in One task with four degradation types, further validating the effectiveness of our approach. For more details on the All in One experimental setup and the comparison results for six types of degradation, please refer to the supplementary.

\begin{table}[!h]
    \centering
    \scriptsize
    \fboxsep0.75pt  
    \setlength\tabcolsep{6pt}
    % \vspace{-2pt}
    \captionsetup{skip=0pt}
    \captionsetup{font=small}
    \caption{Quantitative results of low-light image enhancement.}
    \label{tab:exp:low-light}

    \begin{tabularx}{0.9\linewidth}{Xc*{3}{c}c}
        \toprule
        Method & FS & Params(M)& PSNR&SSIM & LIPIPS \\
        \midrule
        Restormer~\citep{Zamir2021Restormer} & \ding{55}& 26.1 & 21.54 & \cellcolor{pltorange!50}{.843} & \cellcolor{pltyellow!50}{.3608} \\
        LLformer~\citep{wang2023ultra} & \ding{55}& 13.2 & 24.06 & \cellcolor{pltorange!50}{.858} & \cellcolor{pltyellow!50}{.3516} \\
        UHDFour~\citep{Li2023ICLR} & \ding{51}& 17.5 & 26.22 & \cellcolor{pltorange!50}{.900} & \cellcolor{pltyellow!50}{.2390} \\
        UHDformer~\citep{wang2024uhdformer} & \ding{51}& 0.3393  & {27.11} & \cellcolor{pltorange!50}{{.927}} & \cellcolor{pltyellow!50}{{.2240}} \\
        LMAR~\citep{Empowering} & \ding{51}& 1.965 & 26.27 & \cellcolor{pltorange!50}{.919} & \cellcolor{pltyellow!50}{.2248} \\
        UHDDIP \citep{UHDDIP} & \ding{51}&0.81 & 26.74 & \cellcolor{pltorange!50}{.928} & \cellcolor{pltyellow!50}{{.2076}} \\
        DreamUHD~\citep{DreamUHD} & \ding{51}& 1.215 & \textcolor{tabblue}{\textbf{27.72}} & \cellcolor{pltorange!50}{\textcolor{tabblue}{\textbf{.928}}} & \cellcolor{pltyellow!50}{\textcolor{tabblue}{\textbf{.2051}}} \\
        Ours & \ding{51}& 1.008 & \textcolor{tabred}{\textbf{27.94}} & \cellcolor{pltorange!50}{\textcolor{tabred}{\textbf{.934}}} & \cellcolor{pltyellow!50}\textcolor{tabred}{\textbf{{.2041}}} \\
        \bottomrule
        \end{tabularx}
\end{table}

\begin{table}[!h]
    \centering
    \scriptsize
    \fboxsep0.75pt  
    \setlength\tabcolsep{6pt}
    % \vspace{-2pt}
    \captionsetup{skip=0pt}
    \caption{Quantitative results of image dehazing.}
    \label{tab:exp:image dehazing}

    \begin{tabularx}{0.95\linewidth}{Xc*{3}{c}c}
        \toprule
        Method & FS & Params(M)& PSNR& SSIM & LIPIPS \\
        \midrule
        Restormer \citep{Zamir2021Restormer} & \ding{55} & 26.1& 12.72 & \cellcolor{pltorange!50}{.693} & \cellcolor{pltyellow!50}{.4560} \\
        Uformer \citep{Wang_2022_CVPR} & \ding{55} & 20.6& 19.83 & \cellcolor{pltorange!50}{.921} & \cellcolor{pltyellow!50}{.4220} \\
        DehazeFormer\citep{song2023vision} & \ding{55} & 2.5& 15.37 & \cellcolor{pltorange!50}{.737} & \cellcolor{pltyellow!50}{.3998} \\
        TaylorFormer \citep{MB-TaylorFormer} & \ding{55} & 2.7& 20.99 & \cellcolor{pltorange!50}{.919} & \cellcolor{pltyellow!50}{.3124} \\
        UHDformer \citep{wang2024uhdformer} & \ding{51} & 0.3393 & {22.58} & \cellcolor{pltorange!50}{.942} & \cellcolor{pltyellow!50}{{.1188}} \\
        UHDDIP \citep{UHDDIP} & \ding{51}&0.81 & \textcolor{tabblue}{\textbf{24.69}} & \cellcolor{pltorange!50}{\textcolor{tabblue}{\textbf{.952}}} & \cellcolor{pltyellow!50}{\textcolor{tabblue}{\textbf{.1049}}} \\
         DreamUHD \citep{DreamUHD} & \ding{51} & 1.215& 24.36 & \cellcolor{pltorange!50}{.945} & \cellcolor{pltyellow!50}{.1176} \\
        Ours & \ding{51} & 1.008 & \textcolor{tabred}{\textbf{25.37}} & \cellcolor{pltorange!50}{\textcolor{tabred}{\textbf{.955}}} & \cellcolor{pltyellow!50}\textcolor{tabred}{\textbf{{.1044}}} \\
        \bottomrule        
        \end{tabularx}
\end{table}

\begin{table}[!h]
    \centering
    \scriptsize
    \fboxsep0.75pt  
    \setlength\tabcolsep{6pt}
    % \vspace{-2pt}
    \captionsetup{skip=0pt}
    \caption{Quantitative results of image deblurring.}
    \label{tab:exp:image deblurring}

    \begin{tabularx}{0.95\linewidth}{Xc*{3}{c}c}
        \toprule
        Method & FS & Params(M)& PSNR&SSIM & LIPIPS \\
        \midrule
        Restormer \citep{Zamir2021Restormer} & \ding{55} & 26.1& 25.21 & \cellcolor{pltorange!50}{.752} & \cellcolor{pltyellow!50}{.3695} \\
        Uformer \citep{Wang_2022_CVPR} & \ding{55}& 20.6 & 25.26 & \cellcolor{pltorange!50}{.751} & \cellcolor{pltyellow!50}{.3851} \\
        Stripformer \citep{tsai2022stripformer} & \ding{55}& 19.7 & 25.05 & \cellcolor{pltorange!50}{.750} & \cellcolor{pltyellow!50}{.3740} \\
        FFTformer \citep{kong_2023} & \ding{55}& 16.6 & 25.41 & \cellcolor{pltorange!50}{.757} & \cellcolor{pltyellow!50}{.3708} \\
        UHDformer \citep{wang2024uhdformer} & \ding{51}& 0.3393  & {28.82} & \cellcolor{pltorange!50}{.844} & \cellcolor{pltyellow!50}{{.2350}} \\
        UHDDIP \citep{UHDDIP} & \ding{51}&0.81 & \textcolor{tabblue}{\textbf{29.51}} & \cellcolor{pltorange!50}\textcolor{tabblue}{\textbf{.858}} & \cellcolor{pltyellow!50}\textcolor{tabblue}{\textbf{{.2127}}} \\
        DreamUHD~\citep{DreamUHD} & \ding{51}& 1.456 & 29.33 & \cellcolor{pltorange!50}{.852} & \cellcolor{pltyellow!50}{.2222} \\
        Ours & \ding{51}& 1.008 & \textcolor{tabred}{\textbf{29.84}} & \cellcolor{pltorange!50}{\textcolor{tabred}{\textbf{.861}}} & \cellcolor{pltyellow!50}\textcolor{tabred}{\textbf{{.2124}}} \\
        \bottomrule      
        \end{tabularx}
\end{table}

\begin{table}[!h]
    \centering
    \scriptsize
    \fboxsep0.75pt  
    \setlength\tabcolsep{6pt}
    % \vspace{-2pt}
    \captionsetup{skip=0pt}
    \caption{Quantitative results of image demoir´eing.}
    \label{tab:exp:image demoir´eing}

    \begin{tabularx}{0.95\linewidth}{Xc*{3}{c}c}
        \toprule
        Method & FS& Params(M) & PSNR&SSIM & LIPIPS \\
        \midrule
        $\rm FHDe^2Net$\citep{Zamir2021Restormer} & \ding{55}&13.571 & 20.33 & \cellcolor{pltorange!50}{.749} & \cellcolor{pltyellow!50}{.3519} \\
        ESDNet \citep{yu2022towards} & \ding{51}&5.93 & 22.11 & \cellcolor{pltorange!50}{.795} & \cellcolor{pltyellow!50}{.2551} \\
        ESDNet-L \citep{yu2022towards} & \ding{51}&10.62 & 22.42 & \cellcolor{pltorange!50}{.798} & \cellcolor{pltyellow!50}\textcolor{tabred}{\textbf{.2454}} \\
        UHDformer \citep{wang2024uhdformer} & \ding{51}&0.3393 & 21.96 & \cellcolor{pltorange!50}{.833} & \cellcolor{pltyellow!50}{.3854} \\
        UHDDIP \citep{UHDDIP} & \ding{51}&0.81 & {22.06} & \cellcolor{pltorange!50}{.802} & \cellcolor{pltyellow!50}{{.3822}} \\
        DreamUHD \citep{DreamUHD} & \ding{51}&1.456 & \textcolor{tabblue}{\textbf{23.24}} & \cellcolor{pltorange!50}{\textcolor{tabblue}{\textbf{.843}}} & \cellcolor{pltyellow!50}{.3259} \\
        Ours & \ding{51}&1.008 & \textcolor{tabred}{\textbf{23.92}} & \cellcolor{pltorange!50}{\textcolor{tabred}{\textbf{.851}}} & \cellcolor{pltyellow!50}\textcolor{tabblue}{\textbf{{.2842}}} \\
        \bottomrule        
        \end{tabularx}
\end{table}

\begin{table*}[!h]
    \centering
    \scriptsize
    \fboxsep0.75pt
    \setlength\tabcolsep{3pt}
    \caption{\textit{Comparison to state-of-the-art on four degradations.} PSNR (dB, $\uparrow$), \colorbox{pltorange!50}{SSIM ($\uparrow$)}, and \colorbox{pltyellow!50}{LPIPS ($\downarrow$)}, and FS represents full-size 4K image inference. FLOPs are computed for an input size of 256$\times$256. \textcolor{tabred}{\textbf{Best}} and \textcolor{tabblue}{\textbf{second best}} performances are highlighted.}
    \label{tab:exp:4deg}
    \begin{tabularx}{0.9\textwidth}{Xc*{17}{c}}
    \Xhline{1pt}
     \multirow{2}{*}{Method} & \multirow{2}{*}{FS} & \multirow{2}{*}{\centering FLOPs} & \multirow{2}{*}{\centering Params.}  
     & \multicolumn{2}{c}{\textit{Low Light}} & \multicolumn{2}{c}{\textit{Deblurring}} & \multicolumn{2}{c}{\textit{Dehazing}} & \multicolumn{6}{c}{\textit{Denoising}} 
     & \multicolumn{3}{c}{\multirow{2}{*}{\centering Average}} \\
     \cmidrule(lr){5-6} \cmidrule(lr){7-8} \cmidrule(lr){9-10} \cmidrule(lr){11-16}
     &&&& \multicolumn{2}{c}{UHD-LL} & \multicolumn{2}{c}{UHD-blur} & \multicolumn{2}{c}{UHD-haze} & \multicolumn{2}{c}{UHDN\textsubscript{$\sigma$=15}} & \multicolumn{2}{c}{UHDN\textsubscript{$\sigma$=25}} & \multicolumn{2}{c}{UHDN\textsubscript{$\sigma$=50}} \\
     \midrule
        AIRNet~\citep{AirNet} & \ding{55} & 301G & {9M} & 19.24 & \cellcolor{pltorange!50}{.809} & 21.89 &  \cellcolor{pltorange!50}{.757} & 18.37 &  \cellcolor{pltorange!50}{.812} & 21.33 &  \cellcolor{pltorange!50}{.887} & 20.78 &  \cellcolor{pltorange!50}{.784} & 18.79 &  \cellcolor{pltorange!50}{.475} & 20.07 &  \cellcolor{pltorange!50}{.754} & \cellcolor{pltyellow!50}{.2843} \\
        IDR~\citep{zhang2023ingredient} & \ding{55} & 88G & 15.3M & 23.12 & \cellcolor{pltorange!50}{.910} & 24.67 &  \cellcolor{pltorange!50}{.793} & 19.12 &  \cellcolor{pltorange!50}{.768} & 27.48 &  \cellcolor{pltorange!50}{.912} & 25.86 &  \cellcolor{pltorange!50}{.872} & 24.57 &  \cellcolor{pltorange!50}{.654} & 24.14 &  \cellcolor{pltorange!50}{.822} & \cellcolor{pltyellow!50}{.2684} \\
        PromptIR~\citep{potlapalli2023promptir} & \ding{55} & 158G & 33M & 23.44 & \cellcolor{pltorange!50}{.902} & 25.77 & \cellcolor{pltorange!50}{.782} & 19.97 & \cellcolor{pltorange!50}{.727} & 28.43 & \cellcolor{pltorange!50}{.924} & 26.74 & \cellcolor{pltorange!50}{.898} & 23.72 & \cellcolor{pltorange!50}{.584} & 24.68 & \cellcolor{pltorange!50}{.803} & \cellcolor{pltyellow!50}{.2571} \\
        CAPTNet~\citep{gao2023prompt} & \ding{55} & 25G & 24.3M & 23.96 & \cellcolor{pltorange!50}{.920} & 26.11 & \cellcolor{pltorange!50}{.798} & 19.46 & \cellcolor{pltorange!50}{.868} & 25.58 & \cellcolor{pltorange!50}{.865} & 23.24 & \cellcolor{pltorange!50}{.884} & 21.98 & \cellcolor{pltorange!50}{.508} & 23.39 & \cellcolor{pltorange!50}{.809} & \cellcolor{pltyellow!50}{.3466} \\
        
        NDR-Restore~\citep{yao2024neural} & \ding{55} & 196G & 36.9M & 23.84 & \cellcolor{pltorange!50}{.894} & 24.25 & \cellcolor{pltorange!50}{.802} & 20.08 & \cellcolor{pltorange!50}{.892} & 25.62 & \cellcolor{pltorange!50}{.912} & 24.37 & \cellcolor{pltorange!50}{.897} & 22.94 & \cellcolor{pltorange!50}{.669} & 23.52 & \cellcolor{pltorange!50}{.846} & \cellcolor{pltyellow!50}{.3126} \\
        
        Gridformer~\citep{wang2024gridformer} & \ding{55} & 367G & 34M & 23.12 & \cellcolor{pltorange!50}{.898} & 25.82 & \cellcolor{pltorange!50}{.783} & 19.24 & \cellcolor{pltorange!50}{.869} & 36.04 & \cellcolor{pltorange!50}{.937} & 31.72 & \cellcolor{pltorange!50}{.898} & 26.24 & \cellcolor{pltorange!50}{.623} & 27.03 & \cellcolor{pltorange!50}{.836} & \cellcolor{pltyellow!50}{.3754} \\
        
        DiffUIR-L~\citep{zheng2024selective} & \ding{55} & {10G} & 36.2M & 21.56 & \cellcolor{pltorange!50}{.812} & 23.85 & \cellcolor{pltorange!50}{.743} & 18.28 & \cellcolor{pltorange!50}{.864} & \textcolor{tabblue}{\textbf{36.84}} & \cellcolor{pltorange!50}\textcolor{tabblue}{\textbf{{.938}}} & \textcolor{tabblue}{\textbf{32.42}} & \cellcolor{pltorange!50}{.897} & 26.08 & \cellcolor{pltorange!50}{.648} & 26.51 & \cellcolor{pltorange!50}{.818} & \cellcolor{pltyellow!50}\textcolor{tabblue}{\textbf{{.2564}}} \\
        
        Histoformer~\citep{sun2025restoring} & \ding{55} & 91G & 16.6M & 23.22 & \cellcolor{pltorange!50}{.908} & 25.62 & \cellcolor{pltorange!50}{.782} & 19.78 & \cellcolor{pltorange!50}\textcolor{tabblue}{\textbf{{.903}}} & 26.88 & \cellcolor{pltorange!50}{.845} & 25.64 & \cellcolor{pltorange!50}{.874} & 23.13 & \cellcolor{pltorange!50}{.659} & 24.04 & \cellcolor{pltorange!50}{.829} & \cellcolor{pltyellow!50}{.3524} \\

        adaIR~\citep{cui2024adair} & \ding{55} & 147G & 28.7M & 23.57 & \cellcolor{pltorange!50}{.916} & \textcolor{tabblue}{\textbf{26.35}} & \cellcolor{pltorange!50}\textcolor{tabblue}{\textbf{{.801}}} & 18.44 & \cellcolor{pltorange!50}{.901} & 32.84 & \cellcolor{pltorange!50}{.921} & 30.48 & \cellcolor{pltorange!50}\textcolor{tabblue}{\textbf{{.901}}} & \textcolor{tabblue}{\textbf{26.48}} & \cellcolor{pltorange!50}\textcolor{tabblue}{\textbf{{.672}}} & 26.36 & \cellcolor{pltorange!50}\textcolor{tabblue}{\textbf{{.857}}} & \cellcolor{pltyellow!50}{.3429} \\
        
        HAIR~\citep{cao2024hairhypernetworksbasedallinoneimage} & \ding{55} & 41G & 29M & \textcolor{tabblue}{\textbf{25.75}} & \cellcolor{pltorange!50}\textcolor{tabblue}{\textbf{{.922}}} & 25.78 & \cellcolor{pltorange!50}{.798} & \textcolor{tabblue}{\textbf{20.00}} & \cellcolor{pltorange!50}{.894} & 35.54 & \cellcolor{pltorange!50}{.916} & 30.84 & \cellcolor{pltorange!50}{.878} & 26.26 & \cellcolor{pltorange!50}{.657} & {27.36} & \cellcolor{pltorange!50}{.847} & \cellcolor{pltyellow!50}{.2822} \\

        UHDformer~\citep{wang2024uhdformer} & \ding{51} & 3.0G & 0.33M & {23.22} & \cellcolor{pltorange!50}{{.904}} & 25.44 & \cellcolor{pltorange!50}{.782} & {19.48} & \cellcolor{pltorange!50}{.911} & 34.72 & \cellcolor{pltorange!50}{.942} & 30.27 & \cellcolor{pltorange!50}{.881} & 26.08 & \cellcolor{pltorange!50}{.653} & {26.53} & \cellcolor{pltorange!50}{.846} & \cellcolor{pltyellow!50}{.2878} \\

        UHDDIP~\citep{wang2024ultrahighdefinitionrestorationnewbenchmarks} & \ding{51} & 2.2G & 0.81M & {22.32} & \cellcolor{pltorange!50}{{.894}} & 25.68 & \cellcolor{pltorange!50}{.792} & {19.84} & \cellcolor{pltorange!50}{.899} & 36.42 & \cellcolor{pltorange!50}{.922} & 29.33 & \cellcolor{pltorange!50}{.872} & 26.21 & \cellcolor{pltorange!50}{.652} & {26.63} & \cellcolor{pltorange!50}{.839} & \cellcolor{pltyellow!50}{.2878} \\

        DreamUHD~\citep{DreamUHD} & \ding{51} & 4.1G & 1.46M & {24.23} & \cellcolor{pltorange!50}{{.915}} & 25.84 & \cellcolor{pltorange!50}{.808} & {19.92} & \cellcolor{pltorange!50}{.902} & 36.59 & \cellcolor{pltorange!50}{.927} & 31.64 & \cellcolor{pltorange!50}{.886} & 26.38 & \cellcolor{pltorange!50}{.657} & \textcolor{tabblue}{\textbf{27.43}} & \cellcolor{pltorange!50}{.849} & \cellcolor{pltyellow!50}{.2822} \\

        Ours & \ding{51} & {4.0G} & {1.0M} & \textcolor{tabred}{\textbf{25.89}} & \cellcolor{pltorange!50}\textcolor{tabred}{\textbf{.924}} & \textcolor{tabred}{\textbf{26.42}} & \cellcolor{pltorange!50}\textcolor{tabred}{\textbf{.803}} & \textcolor{tabred}{\textbf{20.34}} & \cellcolor{pltorange!50}\textcolor{tabred}{\textbf{.914}} & \textcolor{tabred}{\textbf{37.13}} & \cellcolor{pltorange!50}\textcolor{tabred}{\textbf{.941}} & \textcolor{tabred}{\textbf{32.89}} & \cellcolor{pltorange!50}\textcolor{tabred}{\textbf{.904}} & \textcolor{tabred}{\textbf{26.54}} & \cellcolor{pltorange!50}\textcolor{tabred}{\textbf{.675}} & \textcolor{tabred}{\textbf{28.20}} & \cellcolor{pltorange!50}\textcolor{tabred}{\textbf{.860}} & \cellcolor{pltyellow!50}\textcolor{tabred}{\textbf{.2523}} \\

        % Continue adding other rows as needed
        \Xhline{1pt}
    \end{tabularx}
\end{table*}

\section{Ablation Study}

In this section, we conduct ablation experiments on the UHD image deblurring task using CD²-VAE, CIMF-Net, and LaReNet to validate their effectiveness.
\subsection{Controlled Differential Disentangled VAE}

\begin{figure}[t!]
    \centering
    \includegraphics[width=0.5\textwidth]{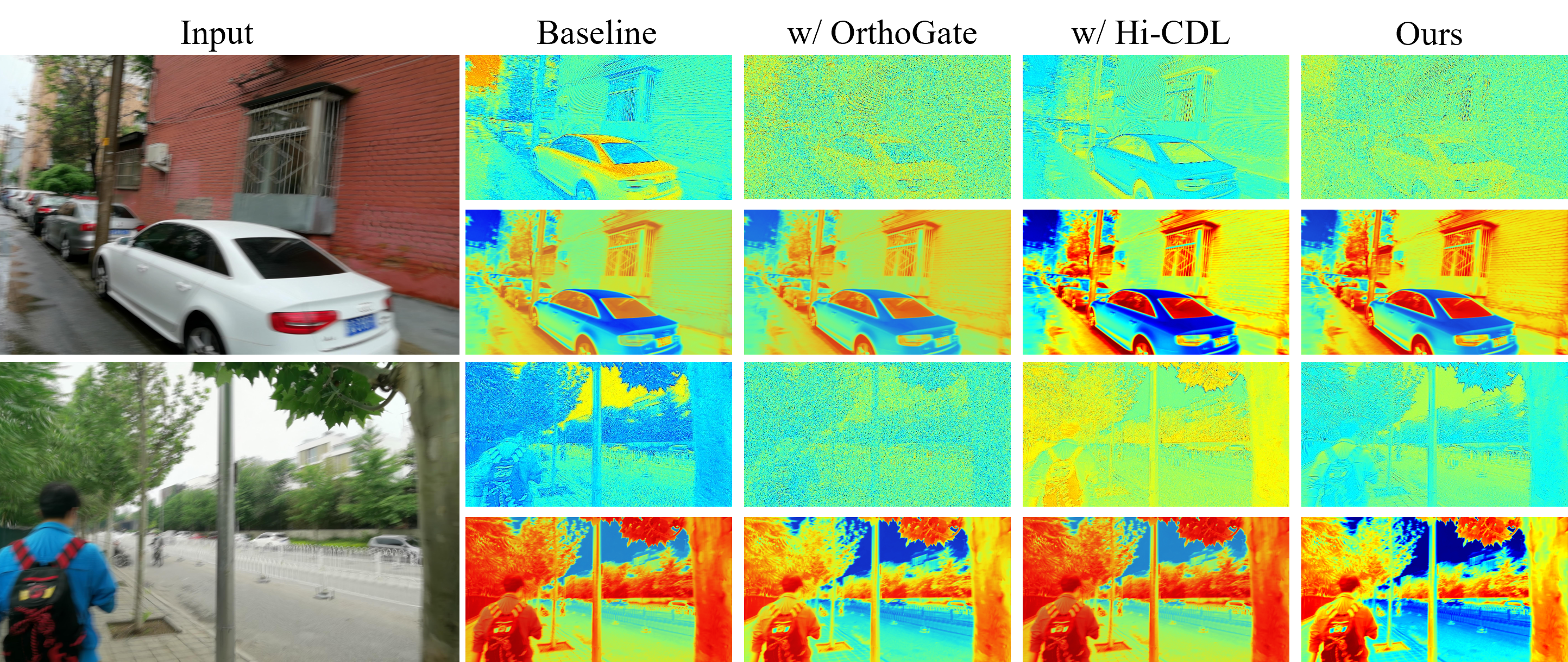} %
    % \captionsetup{skip=0.5pt}
    \caption{Feature visualization of CD²-VAE ablation experiments. The first row shows the encoded latent representation, and the second row shows the discarded information during encoding.}
    \label{fig:ab1}
    % \vspace{-1em}
\end{figure}

\begin{figure}[t!]
    \centering
    \includegraphics[width=0.5\textwidth]{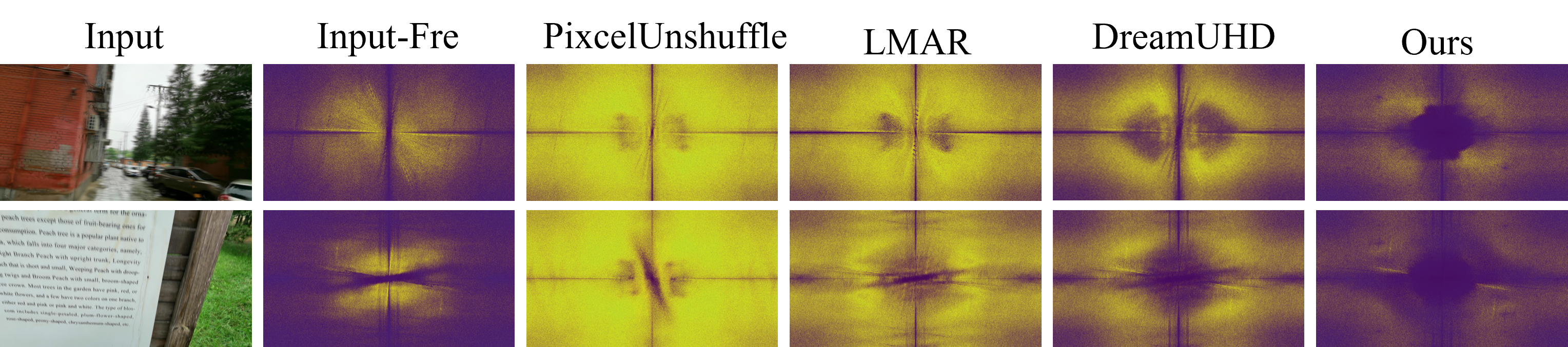} %
        % \captionsetup{skip=0.5pt}
    \caption{Frequency spectrum residual maps of different compression methods compared to GT, with "Input-fre" representing the residual map between input and GT.}
    \label{fig:information loss}
    % \vspace{-1em}
\end{figure}

For the ablation experiments of the proposed CD²-VAE, as shown in ~\cref{tab:ablation_CDVAE}, it can be seen that the two decoupling mechanisms introduced in CD²-VAE, Hi-CDL and OrthoGate, significantly improve performance, with further enhancement achieved through their synergistic effect. We performed feature visualization of the latent and discarded information encoded by the VAE in all ablation experiments, with results shown in ~\cref{fig:ab1}. It can be observed that due to the deep coupling of degraded components and background components in the degraded images, the baseline method indiscriminately compresses all components, leading to uncontrollable information loss during the encoding process. After introducing Hi-CDL and OrthoGate, the information encoding loss becomes more controllable, allowing degradation-dominated features to be encoded as latent , while background-dominated features are separated during the encoding process. OrthoGate tends to suppress the flow of background-dominated information into deeper encoding layers, while Hi-CDL more precisely captures the degradation patterns. The combined effect of both mechanisms results in the optimal decoupling performance.

\begin{table}[!t]
\centering
\scriptsize
\setlength{\tabcolsep}{4pt} % Adjust table column spacing
\captionsetup{skip=0.5pt}
\caption{Ablation Study of the CD²-VAE.}
\begin{tabular}{lcccccc}
\toprule % Top line (thinner)
Method & baseline & w/ Hi-CDL & w/ OrthoGate & Ours \\
\midrule % Middle line (default thickness)
PSNR & 28.89 & 29.48 & 29.32 & \textbf{29.84} \\
SSIM & 0.836 & 0.852 & 0.853 & \textbf{0.861}  \\
\bottomrule % Bottom line (thinner)
\end{tabular}
\label{tab:ablation_CDVAE} 
\end{table}

\subsection{Complex Invertible Multiscale Fusion Network}

In the ablation study of CIMF-Net, as shown in ~\cref{tab:ablation_CIMF}, we conducted comparisons with the following experimental setups: Set1, which is a non-reversible network with a similar number of parameters; Set2, where each scale's background-dominated features are processed by separate INN networks with no interaction between scales; Set3, which builds on Set2 and introduces cross-attention to enable interaction across scales; and Set4, where the real and imaginary parts of the complex numbers in CIMF-Net are concatenated, fused through convolution, and processed by a real-valued network, with additional INN modules to match CIMF-Net's parameter count. The results show that CIMF-Net outperforms Set1 in terms of information preservation capability, thanks to its reversibility. Additionally, compared to Set2, Set3, and Set4, CIMF-Net achieves the best performance by leveraging complex-domain operations that facilitate efficient cross-scale interactions.

\subsection{Latent Restoration Network}

We conducted ablation and comparison experiments on LaReNet using three widely used backbones for image restoration: Restormer~\cite{Zamir2021Restormer}, NAFNet~\cite{chen2022simple}, and SFHformer~\cite{jiang2024fast}. The results show that D²R-UHDNet achieves excellent performance (PSNR) across all backbones and significantly improves the efficiency (Params, FLOPs, Runtime) and performance of each backbone in UHD restoration tasks. This effectively demonstrates the generalizability of the D²R-UHDNet.

\subsection{Analysis of Information Loss}

We visualize the information loss from different downsampling methods as residual maps compared to the GT magnitude spectrum, as shown in ~\cref{fig:information loss} .It can be observed that traditional information compression methods all face varying degrees of significant information loss. Although DreamUHD alleviates high-frequency information loss through a specialized high-frequency compensation mechanism, the loss of mid-frequency information remains unaddressed. It is evident that, due to the controllability in both the information loss and compensation phases, our method exhibits significantly better information retention compared to other methods. This validates the necessity and effectiveness of the image component decoupling strategy.

% \vspace{-5pt} 

\begin{table}[!]
\centering
\scriptsize
\setlength{\tabcolsep}{5pt} % Adjust table column spacing
\captionsetup{skip=0.5pt}
\caption{Ablation Study of CIMF-Net.}
\begin{tabular}{lccccccc}
\toprule % Top line (thinner)
Method & Set1 & Set2 & Set3 & Set4   & Ours \\
\midrule % Middle line (default thickness)
PSNR & 27.86 & 29.26 & 29.52 & 28.88  & \textbf{29.84} \\
SSIM & 0.852 & 0.857 & 0.860 & 0.852  & \textbf{0.861} \\
Params(M)&1.103& 2.212& 4.289.& 1.114& \textbf{1.008}  \\
\bottomrule % Bottom line (thinner)
\end{tabular}
\label{tab:ablation_CIMF} 
\end{table}

\begin{table}[htbp]
\centering
\scriptsize
\setlength{\tabcolsep}{3pt}
\captionsetup{skip=0.5pt}
\caption{Ablation Study of LaReNet.}
\label{tab:performance}
\begin{tabular}{@{}l *{3}{cc} @{}} 
\toprule
\multirow{2}{*}{metric} & 
\multicolumn{2}{c}{Restormer} & 
\multicolumn{2}{c}{NAFNet} & 
\multicolumn{2}{c}{SFHformer} \\
\cmidrule(lr){2-3} \cmidrule(lr){4-5} \cmidrule(lr){6-7}
 & Base & +D²R & Base & +D²R & Base & +D²R \\
\midrule
PSNR (dB)    & 25.21 & 29.78/ \textcolor{red}{{\textbf{+4.57}} }  & 26.06 & 29.88/ \textcolor{red}{{\textbf{+3.82}}}  & 25.67 & 29.84/ \textcolor{red}{{\textbf{+4.17}} } \\
Param (M)   & 26.1  & 3.2/ \textcolor{red}{\textbf{-87}\%\ }   & 29.1  & 1.6/ \textcolor{red}{\textbf{-94}\%\ } & 7.6  & 1.0/ \textcolor{red}{\textbf{-87}\%\ }  \\
FLOPS (G)    & 140.9  & 6.4/ \textcolor{red}{\textbf{-95}\%\ }  & 16.1  & 3.9/ \textcolor{red}{\textbf{-76}\%\ }  & 51.0  & 4.0/ \textcolor{red}{\textbf{-92}\%\ }  \\
Runtime (s)& 8.8  & 0.64/ \textcolor{red}{\textbf{-92}\%\ }   & 4.6  & 0.42/ \textcolor{red}{\textbf{-92}\%\ }   & 5.2  & 0.44/ \textcolor{red}{\textbf{-92}\%\ }   \\
FS           & \ding{55} & \ding{51}\ & \ding{55} & \ding{51} & \ding{55} & \ding{51}\\
\bottomrule
\end{tabular}

\smallskip
\end{table}

\section{Conclusion}

This paper introduces D²R-UHDNet, an efficient and effective framework for UHD image restoration. By decoupling background-dominated and degradation-dominated features and addressing them in a divide-and-conquer manner, the method allows for controllable information loss and targeted information compensation. While maintaining efficiency, it significantly mitigates the information loss typically encountered in traditional UHD restoration methods. The method achieves state-of-the-art results across six experimental settings, including low-light enhancement, dehazing, deblurring, and de-moiréing. The Active Discarding and Targeted Restoration paradigm offers a robust foundation for future advancements in UHD image restoration.
{
    \small
    \bibliographystyle{ieeenat_fullname}
    \bibliography{main}
}

\clearpage
\setcounter{page}{1}
\maketitlesupplementary

\noindent This supplementary document is organized as follows:

\noindent~\cref{sec:visual} presents additional visual results.

\noindent ~\cref{sec:six_degre} presents the results under six degradation experimental settings.

\noindent ~\cref{sec:Experimental Details} provides the construction of the UHD  dataset and details of the experimental setup.

\section{More visual comparison results.}
\label{sec:visual}

% We provide additional visual experimental results, where it can be observed that our method achieves the best visual performance.

\begin{figure*}[t!]
	\centering
	\includegraphics[width=1\textwidth]{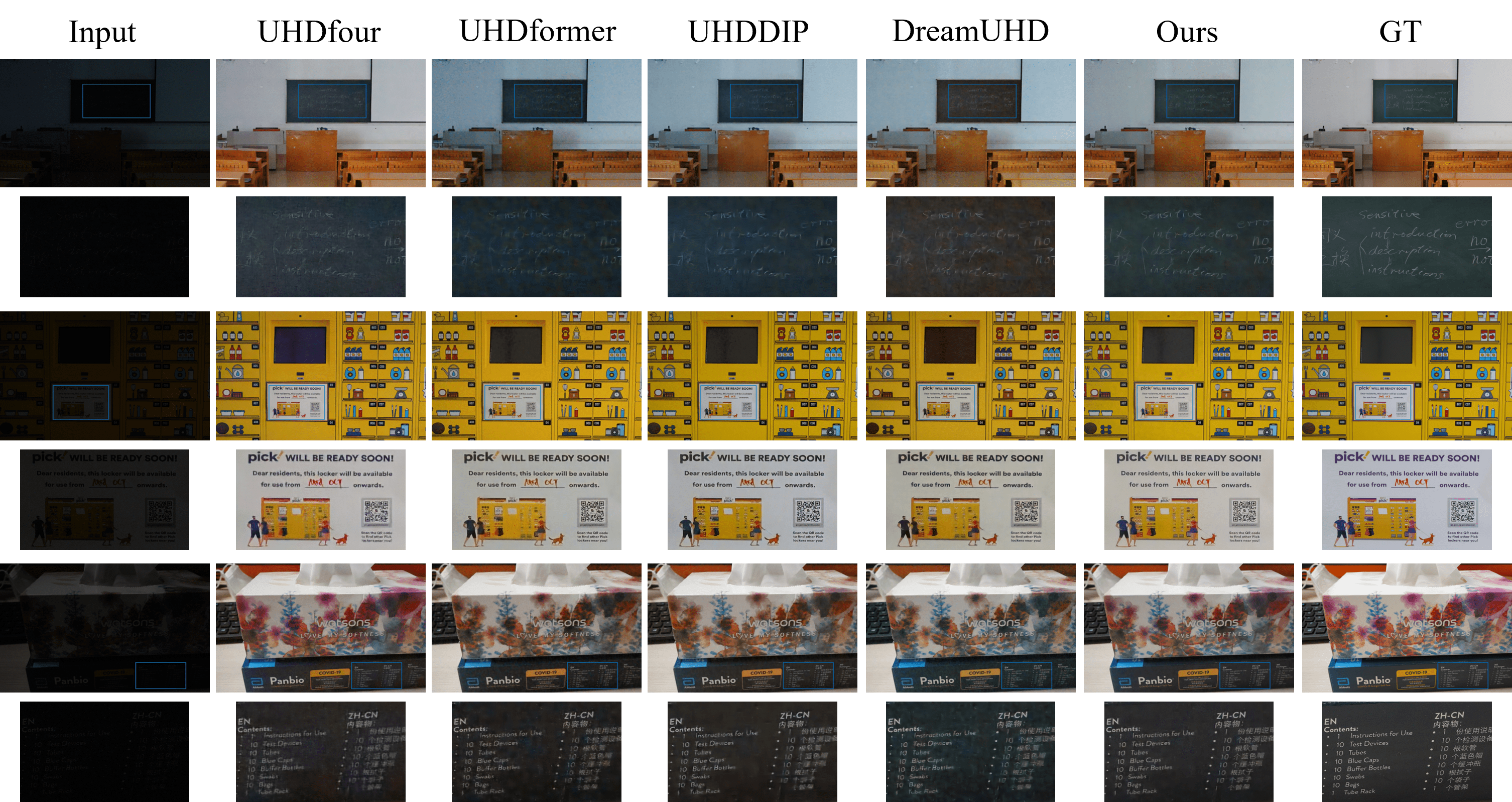} %
        % \captionsetup{skip=0.5pt}
	\caption{Additional visual results for LLIE.}
	\label{fig:visual_LLIE}
\end{figure*}

\begin{figure*}[t!]
	\centering
	\includegraphics[width=1\textwidth]{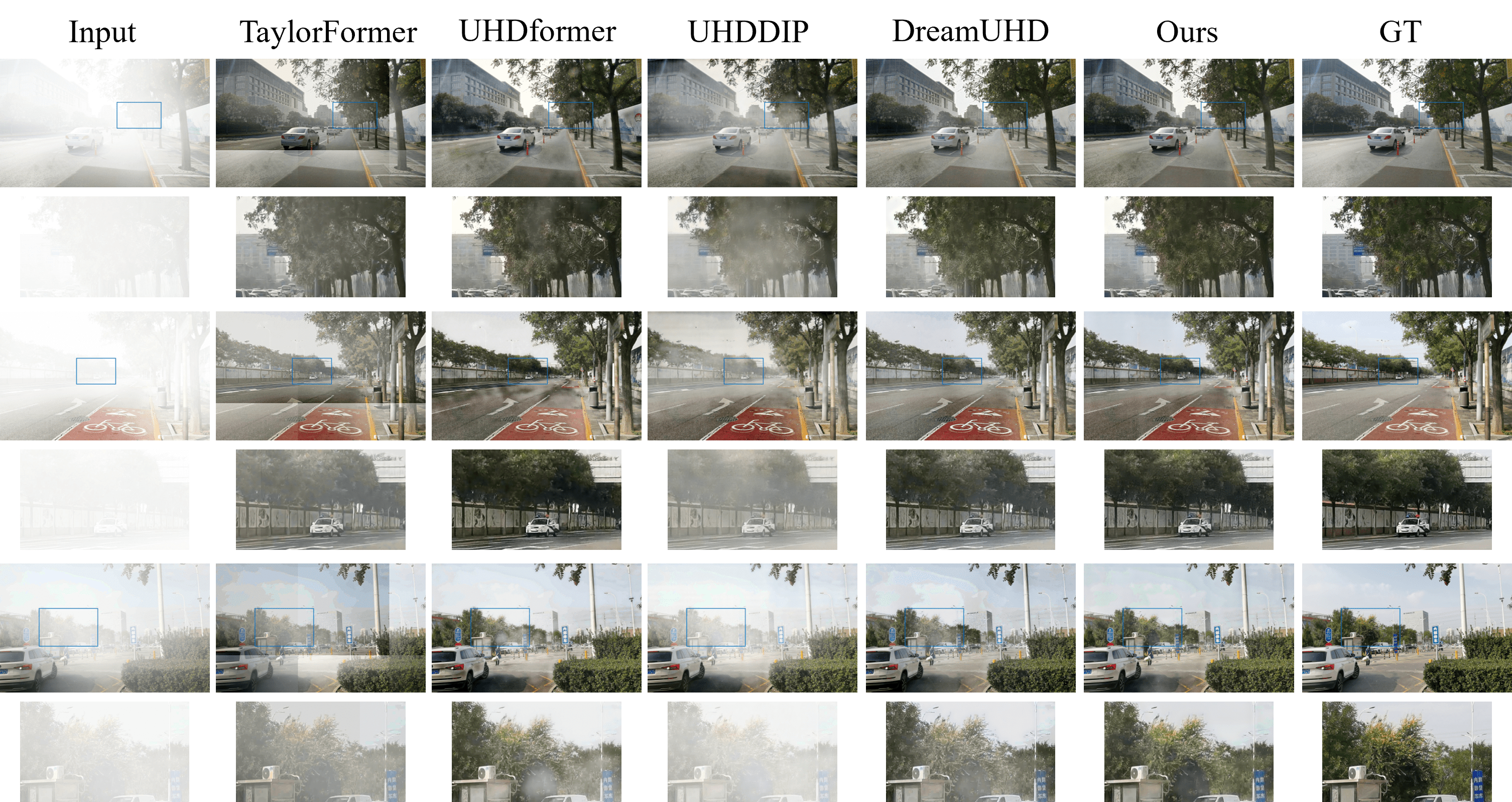} %
        % \captionsetup{skip=0.5pt}
	\caption{Additional visual results for image dehazing.}
	\label{fig:visual_haze}
\end{figure*}

\begin{figure*}[t!]
	\centering
	\includegraphics[width=1\textwidth]{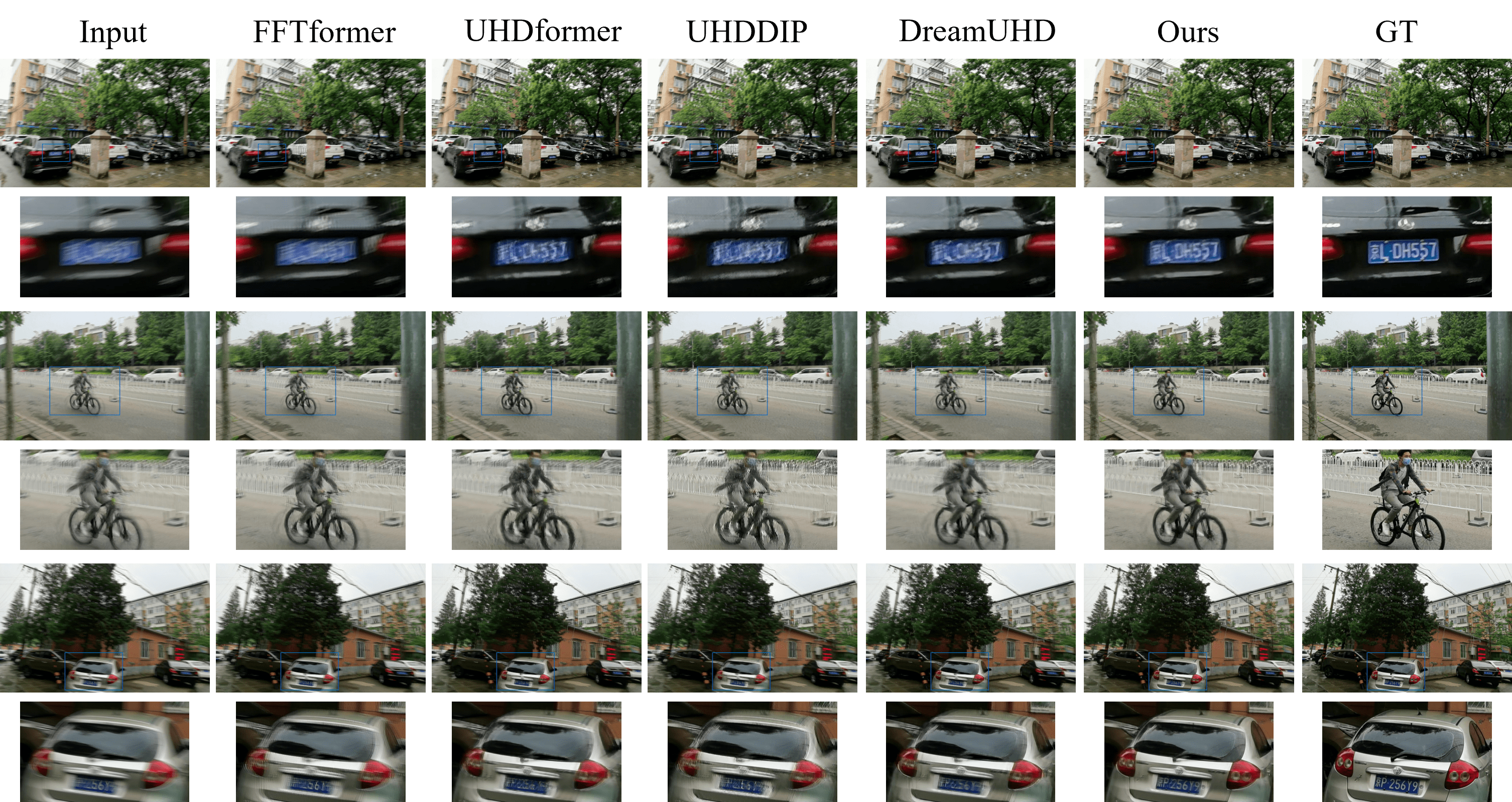} %
        % \captionsetup{skip=0.5pt}
	\caption{Additional visual results for image deblurring.}
	\label{fig:visual_bulr}
\end{figure*}

\begin{figure*}[t!]
	\centering
	\includegraphics[width=1\textwidth]{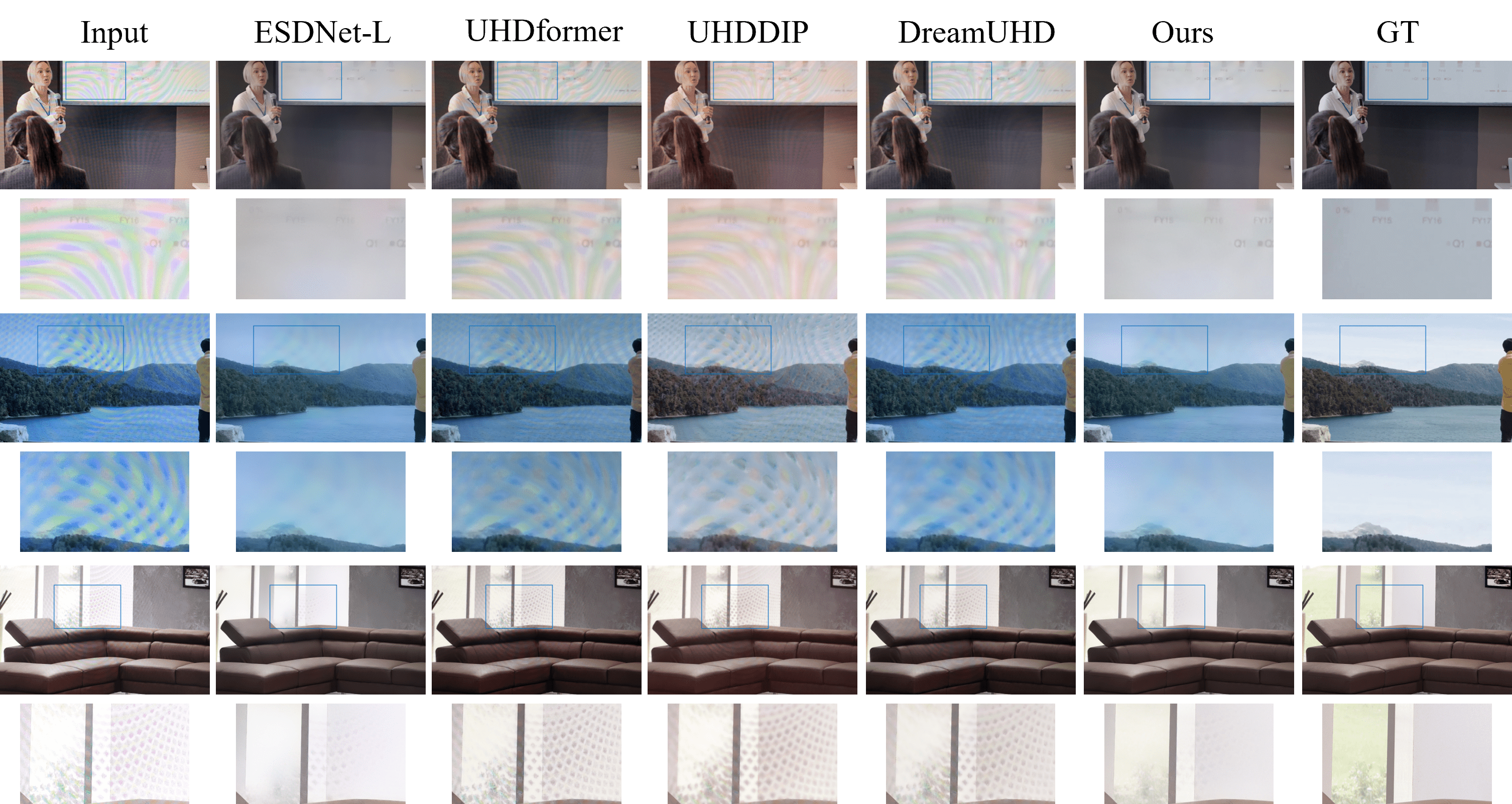} %
        % \captionsetup{skip=0.5pt}
	\caption{Additional visual results for image demoiring.}
	\label{fig:visual_moire}
\end{figure*}

We present additional visual results for low-light image enhancement, image dehazing, image deblurring, and moiré pattern removal in~\cref{fig:visual_LLIE,fig:visual_haze,fig:visual_bulr,fig:visual_moire}. As can be observed, our method achieves minimal degradation artifacts while maintaining the consistency of background information in the images.

\section{The results under six degradation experimental settings }
\label{sec:six_degre}

We further design six types of degraded UHD all-in-one experiments, including low-light enhancement, image deblurring, image dehazing, image denoising, image deraining, and image desnowing. The experimental results are shown in~\cref{tab:exp:6deg}. Our method significantly outperforms both traditional all-in-one approaches and UHD restoration methods. By balancing efficiency and performance, we validate the effectiveness of our method.

\section{Experimental Details}
\label{sec:Experimental Details}
\subsection{Datasets}

The various UHD degradation scenarios in this paper are based on UHD-LL~\cite{Li2023ICLR}, UHD-blur~\cite{deng2021multi}, UHD-haze~\cite{uhdhaze}, UHD-rain~\cite{chen2024ultrahighdefinitionimagederainingbenchmark}, and UHD-snow~\cite{wang2024ultrahighdefinitionrestorationnewbenchmarks}. For UHD denoising, 4k images from~\cite{zhang2021benchmarking} are used as the background.The distributions of the training and testing sets for all datasets are shown in ~\cref{tab:dataset_details}. 

\begin{table*}[!]
\centering
\small
\setlength{\tabcolsep}{4pt} 
% \captionsetup{skip=0.5pt}
\caption{Dataset details and corresponding tasks.}
\begin{tabular}{lccc}
\toprule % 
\textbf{Dataset} & \textbf{Training samples} & \textbf{Testing samples} & \textbf{Task} \\
\midrule % 
UHD-Snow & 2,000 & 200 & Desnowing \\
UHD-Blur & 1,964 & 300 & Deblurring \\
UHD-Rain & 2,000 & 500 & Deraining \\
UHD-LL   & 2,000 & 115 & LLIE \\
UHD-Haze   & 2,290 & 231 & Dehazing \\
UHD-Noise   & 2,000 & 500 & Denoising \\

\bottomrule 
\end{tabular}
\label{tab:dataset_details}
\end{table*}

\subsection{Implementation Details}

The number of encoder and decoder layers is set to 3, the number of modules in the latent image restoration network is set to 6, and the number of Glow modules in CIMF-Net is set to 3.

For the first stage, we train Clean-VAE on the image reconstruction task. The initial learning rate is set to \(5 \times 10^{-4}\), gradually reduced to \(1 \times 10^{-7}\) using cosine annealing. The batch size is set to 16, and the images are randomly cropped to \(256 \times 256\).

In the second stage, we train CD²-VAE based on paired degraded-clean image inputs for feature disentanglement training. On one hand, the degraded latent extracted from the input is combined with the clean latent extracted by the Clean-VAE encoder and input into the degraded decoder for the reconstruction of the degraded image. On the other hand, the disentangled background features are input into Clean-VAE’s decoder for the reconstruction of the clean image. The initial learning rate is set to \(5 \times 10^{-4}\), gradually reduced to \(1 \times 10^{-7}\) using cosine annealing. The batch size is set to 12, and the images are randomly cropped to \(256 \times 256\).

For the third stage, we train D²R-UHDNet on the image restoration task, keeping the parameters of CD²-VAE frozen. We fine-tune the parameters of the LaReNet and CIMF-Net. The initial learning rate is set to \(4 \times 10^{-4}\), gradually reduced to \(1 \times 10^{-7}\) using cosine annealing. The batch size is set to 6, and the images are randomly cropped to \(512 \times 512\).

\begin{table*}[!h]
    \centering
    \scriptsize
    \fboxsep0.75pt
    \setlength\tabcolsep{3pt}
    \caption{\textit{Comparison to state-of-the-art on three degradations.} PSNR (dB, $\uparrow$), \colorbox{pltorange!50}{SSIM ($\uparrow$)}, \colorbox{pltyellow!50}{LPIPS ($\downarrow$)} and FS represents full-size 4K image inference. FLOPs are computed for an input size of 256$\times$256. \textcolor{tabred}{\textbf{Best}} and \textcolor{tabblue}{\textbf{second best}} performances are highlighted.}
    \label{tab:exp:6deg}
    \begin{tabularx}{0.9\textwidth}{Xc*{17}{c}}
    \Xhline{1pt}
     \multirow{2}{*}{Method} & \multirow{2}{*}{FS} & \multirow{2}{*}{\centering FLOPs} & \multirow{2}{*}{\centering Params.}  
     & \multicolumn{2}{c}{\textit{Low Light}} & \multicolumn{2}{c}{\textit{Deblurring}} & \multicolumn{2}{c}{\textit{Dehazing}} & \multicolumn{2}{c}{\textit{Denoising}} 
     &\multicolumn{2}{c}{\textit{Deraining}} & \multicolumn{2}{c}{\textit{Desnowing}} & \multicolumn{3}{c}{\multirow{2}{*}{\centering Average}} \\
     \cmidrule(lr){5-6} \cmidrule(lr){7-8} \cmidrule(lr){9-10} \cmidrule(lr){11-12} \cmidrule(lr){13-14} \cmidrule(lr){15-16}
     &&&& \multicolumn{2}{c}{UHD-LL} & \multicolumn{2}{c}{UHD-blur} & \multicolumn{2}{c}{UHD-haze} & \multicolumn{2}{c}{UHDN\textsubscript{$\sigma$=50}} & \multicolumn{2}{c}{UHD-rain} & \multicolumn{2}{c}{UHD-snow} \\
     \midrule
        AIRNet~\citep{AirNet} & \ding{55} & 301G & {9M} & 22.68 & \cellcolor{pltorange!50}{.887} & 23.52 &  \cellcolor{pltorange!50}{.876} & 18.24 & \cellcolor{pltorange!50}{.846} & 22.38 &  \cellcolor{pltorange!50}{.876} & 26.35 &  \cellcolor{pltorange!50}{.876} & 27.38 &  \cellcolor{pltorange!50}{.924} & 23.43 &  \cellcolor{pltorange!50}{.874} & \cellcolor{pltyellow!50}{.1861} \\

        IDR~\citep{zhang2023ingredient} & \ding{55} & 88G & 15.3M & 24.33 & \cellcolor{pltorange!50}{.915} & 25.64 &  \cellcolor{pltorange!50}{.788} & 18.68 &  \cellcolor{pltorange!50}{.879} & 29.64 &  \cellcolor{pltorange!50}{.906} & 28.82 &  \cellcolor{pltorange!50}{.906} & 30.48 &  \cellcolor{pltorange!50}{.945} & 26.27 &  \cellcolor{pltorange!50}{.890} & \cellcolor{pltyellow!50}{.1912} \\

        PromptIR~\citep{potlapalli2023promptir} & \ding{55} & 158G & 33M& 23.3 & \cellcolor{pltorange!50}{.911} & 26.48 & \cellcolor{pltorange!50}\textcolor{tabblue}{\textbf{{.805}}} & 20.14 & \cellcolor{pltorange!50}{.901} & 24.88 & \cellcolor{pltorange!50}{.835} & 28.89 & \cellcolor{pltorange!50}{.897} & 30.78 & \cellcolor{pltorange!50}{.966} & 25.74 &  \cellcolor{pltorange!50}{.886} & \cellcolor{pltyellow!50}{.2155} \\

        CAPTNet~\citep{gao2023prompt} & \ding{55} & 25G & 24.3M & 24.97 & \cellcolor{pltorange!50}\textcolor{tabblue}{\textbf{{.921}}} & \textcolor{tabblue}{\textbf{26.32}} &  \cellcolor{pltorange!50}{.796} & \textcolor{tabblue}{\textbf{20.32}} &  \cellcolor{pltorange!50}{.903} & 21.64 &  \cellcolor{pltorange!50}{.569} & \textcolor{tabblue}{\textbf{29.34}} &  \cellcolor{pltorange!50}\textcolor{tabblue}{\textbf{{.908}}} & \textcolor{tabblue}{\textbf{32.21}} &  \cellcolor{pltorange!50}\textcolor{tabblue}{\textbf{{.974}}} & 25.80 &  \cellcolor{pltorange!50}{.845} & \cellcolor{pltyellow!50}{.2861} \\
        
        NDR-Restore~\citep{yao2024neural} & \ding{55} & 196G & 36.9M & 25.12 & \cellcolor{pltorange!50}{.885} & 25.64 &  \cellcolor{pltorange!50}{.791} & 19.21 &  \cellcolor{pltorange!50}{.896} & 31.44 &  {{\cellcolor{pltorange!50}{.915}}} & 29.24 &   {{\cellcolor{pltorange!50}{.897}}} & 28.41 &   {{\cellcolor{pltorange!50}{.948}}} & 26.51 &  \cellcolor{pltorange!50}{.889} & \cellcolor{pltyellow!50}{.3108} \\

        Gridformer~\citep{wang2024gridformer} & \ding{55} & 367G & 34M & 23.92 & \cellcolor{pltorange!50}{.898} & 25.68 &  \cellcolor{pltorange!50}{.782} & 18.87 &  \cellcolor{pltorange!50}{.889} & 32.86 &  \cellcolor{pltorange!50}{.915} & 29.37 &  \cellcolor{pltorange!50}{.904} & 28.24 &  \cellcolor{pltorange!50}{.942} & 26.49&  \cellcolor{pltorange!50}{.895} & \cellcolor{pltyellow!50}{.2321} \\
        
        DiffUIR-L~\citep{zheng2024selective} & \ding{55} & {10G} & 36.2M & 22.64 & \cellcolor{pltorange!50}{.902} &  {{25.08}} &   \cellcolor{pltorange!50}{.785} &  {{18.62}} &   {{\cellcolor{pltorange!50}{.889}}} & \textcolor{tabblue}{\textbf{{33.25}}} &  {\cellcolor{pltorange!50}\textcolor{tabblue}{\textbf{{.928}}}} & {{27.89}} &  {\cellcolor{pltorange!50}{.886}} & {{27.36}} &  {\cellcolor{pltorange!50}{.945}} &  {{25.81}} &   {\cellcolor{pltorange!50}{.889}} & \cellcolor{pltyellow!50}\textcolor{tabblue}{\textbf{{.1844}}} \\
        
        Histoformer~\citep{sun2025restoring} & \ding{55} & 91G & 16.6M & 25.73 & \cellcolor{pltorange!50}{.915} &  {{26.55}} &   \cellcolor{pltorange!50}{.796} &  {{18.73}} &   {{\cellcolor{pltorange!50}{.897}}} & {33.05} &  {\cellcolor{pltorange!50}{.924}} & {{27.96}} &  {\cellcolor{pltorange!50}{.884}} & {{27.56}} &  {\cellcolor{pltorange!50}{.971}} &  {{{26.59}} }&   {\cellcolor{pltorange!50}{.898}} & \cellcolor{pltyellow!50}{.1855} \\
        
        adaIR~\citep{cui2024adair} & \ding{55} & 147G & 28.7M & 23.84 & \cellcolor{pltorange!50}{.918} &  {{26.86}} &   \cellcolor{pltorange!50}{.803} &  {{19.34}} &   {{\cellcolor{pltorange!50}\textcolor{tabblue}{\textbf{{.910}}}}} & {32.46} &  {\cellcolor{pltorange!50}{.923}} & {{28.18}} &  {\cellcolor{pltorange!50}{{.901}}} & {{27.72}} &  {\cellcolor{pltorange!50}{.953}} &  {{26.40}} &   {\cellcolor{pltorange!50}\textcolor{tabblue}{\textbf{.901}}} & \cellcolor{pltyellow!50}{.2492} \\

        HAIR~\citep{cao2024hairhypernetworksbasedallinoneimage} & \ding{55} & 41G & 29M & \textcolor{tabblue}{\textbf{25.22}} & \cellcolor{pltorange!50}{.897} &  {{24.77}} &   \cellcolor{pltorange!50}{.799} &  {{18.75}} &   {{\cellcolor{pltorange!50}{.883}}} & {32.50} &  {\cellcolor{pltorange!50}{.915}} & {{28.76}} &  {\cellcolor{pltorange!50}{.893}} & {{27.89}} &  {\cellcolor{pltorange!50}{.968}} &  {{26.31}} &   {\cellcolor{pltorange!50}{.892}} & \cellcolor{pltyellow!50}{.2607} \\

        UHDformer~\citep{wang2024uhdformer} & \ding{51} & 3.0G & 0.33M  & 22.87 & \cellcolor{pltorange!50}{.891} & {24.68} & \cellcolor{pltorange!50}{.792} & 20.02 & \cellcolor{pltorange!50}{.888} & 27.23 & \cellcolor{pltorange!50}{.892} & 28.32 & \cellcolor{pltorange!50}{.953} & 
        {28.24} & \cellcolor{pltorange!50}{{.882}}&
        {25.23} & \cellcolor{pltorange!50}{.883} & \cellcolor{pltyellow!50}{.2012} \\

        UHDDIP~\citep{wang2024ultrahighdefinitionrestorationnewbenchmarks} & \ding{51} & 2.2G & 0.81M & {24.56} & \cellcolor{pltorange!50}{{.887}} & 24.26 & \cellcolor{pltorange!50}{.794} & {19.68} & \cellcolor{pltorange!50}{.872} & 28.12 & \cellcolor{pltorange!50}{.889} & 28.78 & \cellcolor{pltorange!50}{.942} & 28.07 & \cellcolor{pltorange!50}{.893} & {25.58} & \cellcolor{pltorange!50}{.880} & \cellcolor{pltyellow!50}{.2278} \\

        DreamUHD~\citep{DreamUHD} & \ding{51} & 4.1G & 1.46M & {25.12} & \cellcolor{pltorange!50}{{.901}} & 25.82 & \cellcolor{pltorange!50}{.796} & {20.21} & \cellcolor{pltorange!50}{.908} & 29.08 & \cellcolor{pltorange!50}{.901} & 30.42 & \cellcolor{pltorange!50}{.950} & 32.12 & \cellcolor{pltorange!50}{.914} & \textcolor{tabblue}{\textbf{27.13}} & \cellcolor{pltorange!50}{.895} & \cellcolor{pltyellow!50}{.1998} \\

        Ours & \ding{51} & {4G} & {1.0M} & \textcolor{tabred}{\textbf{26.14}} & \textcolor{tabred}{\textbf{\cellcolor{pltorange!50}\textcolor{tabred}{{\textbf{.916}}}}} & \textcolor{tabred}{\textbf{26.87}} & \textcolor{tabred}{\textbf{\cellcolor{pltorange!50}{.799}}} & \textcolor{tabred}{\textbf{20.38}} & \textcolor{tabred}{\textbf{\cellcolor{pltorange!50}{.911}}} & \textcolor{tabred}{\textbf{29.64}} & \textcolor{tabred}{\textbf{\cellcolor{pltorange!50}{.912}}} & \textcolor{tabred}{\textbf{32.28}} & \textcolor{tabred}{\textbf{\cellcolor{pltorange!50}{.968}}} & \textcolor{tabred}{\textbf{33.32}} & \textcolor{tabred}{\textbf{\cellcolor{pltorange!50}{.929}}} & \textcolor{tabred}{\textbf{28.11}} & \textcolor{tabred}{\textbf{\cellcolor{pltorange!50}{.906}}} & \textcolor{tabred}{\textbf{\cellcolor{pltyellow!50}{.1842}}} \\
        
        % Continue adding other rows as needed
        \Xhline{1pt}
    \end{tabularx}
    
\end{table*}

\subsection{Training procedure}
In the first phase, Clean-VAE is trained for the image reconstruction task using clean images. The clean input image is denoted as \({I}_h\), and the corresponding reconstructed image is represented as \({I}_{r1}\). 

The loss function for a standard Variational Autoencoder (VAE) comprises two key terms: the reconstruction loss and the KL divergence loss. The reconstruction loss quantifies the difference between the decoder's output and the original input, ensuring the output remains consistent with the input image \cite{yu2020tutorialvaesbayesrule}. The KL divergence loss regularizes the latent space by encouraging the posterior distribution to align with the prior distribution. This regularization improves the model's ability to generate consistent and continuous reconstructions from similar inputs \cite{zhou2018variational}.

We adopt this approach, where the reconstruction loss and KL divergence loss are defined as follows:
\begin{equation}
    \begin{split}
         \mathcal{L}_{\text{rec}_{1}} &= \frac{1}{N} \sum_{i=1}^{N} \|{I}_{\text{r1}}^{(i)} - {I}_{\text{h}}^{(i)}\|_1, \\
    \mathcal{L}_{\text{KL}} &= D_{\text{KL}}(q({z}|{I}) \| p({z})), 
    \end{split}
\end{equation}
where \(q({z}|{I}_h)\) represents the approximate posterior distribution of the latent variable \({z}\) given the input image \({I}_h\), and \(p({z})\) is the prior distribution, usually chosen as a standard Gaussian distribution \(\mathcal{N}(0, {I})\). The KL divergence \(D_{\text{KL}}\) measures the discrepancy between the posterior distribution \(q({z}|{I}_h)\) and the prior distribution \(p({z})\).

Additionally, we enforce frequency domain consistency in the reconstruction results using the FFT loss, which is defined as:
\begin{equation}
    \mathcal{L}_{\text{FFT}_{1}} = \frac{1}{N} \sum_{i=1}^{N} \| \text{FFT}({I}_{\text{rec}}^{(i)}) - \text{FFT}({I}_{\text{h}}^{(i)}) \|_1.
\end{equation}

In the second phase, the input degraded image is denoted as \({I}_d\), which corresponds to the clean image \({I}_{gt}\). After feature disentanglement learning with CD²-VAE, the degraded image \({I}_{d_{rec}}\) and the clean image \({I}_{gt_{rec}}\) are reconstructed. The same reconstruction loss and frequency loss from the first phase are applied, and they are expressed as follows:

\begin{equation}
    \begin{split}
        \mathcal{L}_{\text{rec}_{2}} &= \frac{1}{N} \sum_{i=1}^{N} \|{I}_{\text{d}_\text{rec}}^{(i)} - {I}_{\text{d}}^{(i)}\|_1+ \|{I}_{\text{gt}_\text{rec}}^{(i)} - {I}_{\text{gt}}^{(i)}\|_1 \\
        \mathcal{L}_{\text{FFT}_{2}} &= \frac{1}{N} \sum_{i=1}^{N} \| \text{FFT}({I}_{\text{d}_\text{rec}}^{(i)}) - \text{FFT}({I}_{\text{d}}^{(i)}) \|_1 \\
        &\quad  \quad  \quad          + \|\text{FFT}({I}_{\text{gt}_\text{rec}}^{(i)}) - \text{FFT}({I}_{\text{gt}}^{(i)}) \|_1
    \end{split}
\end{equation}

In the third phase, the parameters of CD²-VAE are frozen, and D²R-UHDNet takes over the image restoration task. The input consists only of the degraded image \({I}_d\), and the output of this restoration process is the restored image, denoted as \({I}_{r}\). The loss function is constructed by comparing the restored image \({I}_{r}\) with the clean ground truth image \({I}_{gt}\), as follows:

\begin{equation}
    \begin{split}
        \mathcal{L}_{\text{rec}} &= \frac{1}{N} \sum_{i=1}^{N} \|{I}_{r}^{(i)} - {I}_{gt}^{(i)}\|_1 ,  \\
        \mathcal{L}_{\text{FFT}} &= \frac{1}{N} \sum_{i=1}^{N} \| \text{FFT}({I}_{r}^{(i)}) - \text{FFT}({I}_{gt}^{(i)}) \|_1.
    \end{split}
\end{equation}

\end{document}